%% file: arxiv.tex
\documentclass[10pt,twocolumn,letterpaper]{article}

\usepackage[pagenumbers]{cvpr} 

\usepackage{algorithm}
\usepackage{algorithmic}
\usepackage{epsfig}
\usepackage{graphicx}
\usepackage{amsmath}
\usepackage{amssymb}
\usepackage[table]{xcolor} 
\usepackage{array}         
\usepackage{booktabs}
\usepackage{tikz}
\usepackage{tikz}
\usepackage{xcolor} 
\usepackage{multirow}
\usepackage{comment}
\usepackage{lineno}
\usepackage{pifont}
\usepackage{listings}
\usepackage{cuted}
\definecolor{c1}{rgb}{0.435, 0.584, 0.933}
\definecolor{c2}{rgb}{0.533, 0.890, 0.949}
\definecolor{c3}{rgb}{0.145, 0.231, 0.573}
\definecolor{c4}{rgb}{0.290, 0.125, 0.675}
\definecolor{c5}{rgb}{0.376, 0.318, 0.949}
\definecolor{c6}{rgb}{0.917, 0.235, 0.192}
\definecolor{c7}{rgb}{0.917, 0.259, 0.765}
\definecolor{c8}{rgb}{0.541, 0.165, 0.353}
\definecolor{c9}{rgb}{0.917, 0.200, 0.964}
\definecolor{c10}{rgb}{0.945, 0.612, 0.976}
\definecolor{c11}{rgb}{0.267, 0.031, 0.282}
\definecolor{c12}{rgb}{0.627, 0.125, 0.298}
\definecolor{c13}{rgb}{0.961, 0.792, 0.271}
\definecolor{c14}{rgb}{0.933, 0.502, 0.271}
\definecolor{c15}{rgb}{0.310, 0.678, 0.196}
\definecolor{c16}{rgb}{0.494, 0.255, 0.086}
\definecolor{c17}{rgb}{0.671, 0.933, 0.412}
\definecolor{c18}{rgb}{0.988, 0.945, 0.631}
\definecolor{c19}{rgb}{0.914, 0.200, 0.137}
\definecolor{c20}{rgb}{0.980, 0.588, 0.000}
\definecolor{c21}{rgb}{0.196, 1.000, 1.000}
\definecolor{barriercol}{RGB}{169,169,169}
\definecolor{bicyclecol}{RGB}{215,25,28}
\definecolor{buscol}{RGB}{253,141,60}
\definecolor{carcol}{RGB}{255,222,0}
\definecolor{convcol}{RGB}{240,90,20}
\definecolor{motorcyclecol}{RGB}{227,26,28}
\definecolor{pedestriancol}{RGB}{31,120,180}
\definecolor{trafficconecol}{RGB}{0,102,102}
\definecolor{trailercol}{RGB}{255,127,0}
\definecolor{truckcol}{RGB}{235,138,98}
\definecolor{drivecol}{RGB}{48,188,172}
\definecolor{sidewalkcol}{RGB}{100,0,100}
\definecolor{terraincol}{RGB}{140,200,100}
\definecolor{manmadecol}{RGB}{200,180,120}
\definecolor{vegetationcol}{RGB}{0,153,0}
\definecolor{otherflatcol}{RGB}{128,128,128}
\definecolor{othercol}{RGB}{0,0,0}

\definecolor{cvprblue}{rgb}{0.21,0.49,0.74}
\usepackage[pagebackref,breaklinks,colorlinks,allcolors=cvprblue]{hyperref}

\title{OccLE: Label-Efficient 3D Semantic Occupancy Prediction}

\author{
Naiyu Fang$^{1,4,5}$, 
Zheyuan Zhou$^{2}$, 
Fayao Liu$^{3}$, 
Xulei Yang$^{3}$,
Jiacheng Wei$^{4}$, \\
Lemiao Qiu$^{2}$, 
Hongsheng Li$^{5}$,
Guosheng Lin$^{1,4}$\thanks{\footnotesize Corresponding author. G. Lin. (e-mail: \texttt{gslin@ntu.edu.sg})} \\
$^{1}$NTU, S-Lab\quad 
$^{2}$ZJU\quad 
$^{3}$A*STAR\quad 
$^{4}$NTU, CCDS\quad 
$^{5}$CUHK, MMLab \\
E-mail: \texttt{naiyufang@cuhk.edu.hk, gslin@ntu.edu.sg}
}

\begin{document}

\maketitle
\begin{abstract}
3D semantic occupancy prediction offers an intuitive and efficient scene understanding and has attracted significant interest in autonomous driving perception. Existing approaches either rely on full supervision, which demands costly voxel-level annotations, or on self-supervision, which provides limited guidance and yields suboptimal performance. To address these challenges, we propose \textbf{OccLE}, a \textbf{L}abel-\textbf{E}fficient 3D Semantic \textbf{Occ}upancy Prediction that takes images and LiDAR as inputs and maintains high performance with limited voxel annotations. Our intuition is to decouple the semantic and geometric learning tasks and then fuse the learned feature grids from both tasks for the final semantic occupancy prediction. Therefore, the semantic branch distills 2D foundation model to provide aligned pseudo labels for 2D and 3D semantic learning. The geometric branch integrates image and LiDAR inputs in cross-plane synergy based on their inherency, employing semi-supervision to enhance geometry learning. We fuse semantic-geometric feature grids through Dual Mamba and incorporate a scatter-accumulated projection to supervise unannotated prediction with aligned pseudo labels. Experiments show that OccLE achieves competitive performance with only 10\% of voxel annotations on the SemanticKITTI and Occ3D-nuScenes datasets. The code will be publicly released on GitHub. The code will be publicly released on \url{https://github.com/NerdFNY/OccLE}.
\end{abstract}

\section{Introduction}

\label{sec1}
3D perception task is foundational cornerstone for autonomous driving systems. Among various perception methods, 3D semantic occupancy prediction \cite{song2017semantic,li2022modality} has garnered significant attention for providing intuitive and efficient scene understanding for downstream tasks. This task estimates the occupancy status and semantic label of each voxel in a 3D grid, given input from 2D images, LiDAR, or a combination of both.

Previous studies \cite{li2023voxformer,ming2024occfusion} proposed numerous supervised learning paradigms with diverse modal inputs. A notable challenge is that their high performance relies heavily on extensive voxel annotations, which are both costly and labor-intensive. The cubic complexity of the voxel grid leads to a substantial workload during manual annotation. Even with pre-annotated 3D labels generated using auto-labeling assistants, one hour of human labor just only annotates 10 frames \cite{pan2024renderocc,huang2024selfocc}. This issue limits the scalability and robustness of supervised methods in real-world deployment.

\begin{figure}[!t] 
    \centering
    \includegraphics[width=\linewidth]{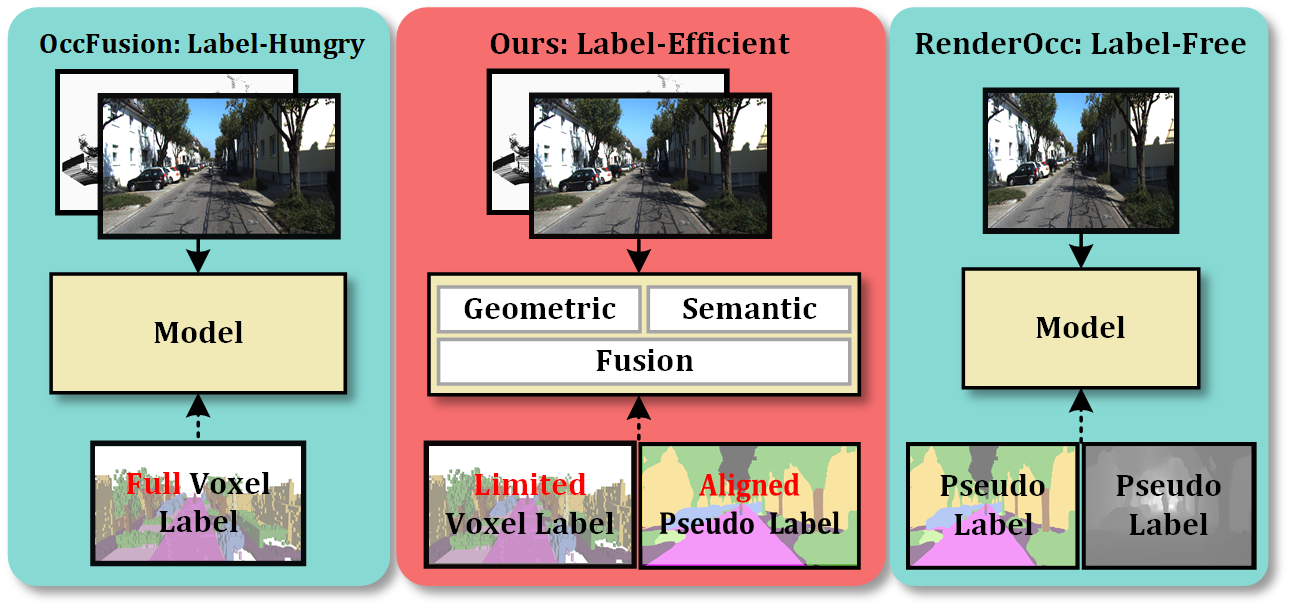} 
    \caption{Label-efficient 3D semantic occupancy prediction aims to achieve high performance using limited voxel annotations and aligned pseudo label. We propose OccLE, a novel learning paradigm that decouples semantic and geometric learning and fuse their feature grids for the final prediction.}
\label{fig1}
\end{figure}

Recent advancements in 3D semantic occupancy prediction have shifted towards a label-free paradigm. These self-supervised methods leverage existing vision foundation models to generate various pseudo labels for auxiliary supervision, such as image semantic segmentation \cite{ ravi2024sam}, depth information \cite{ yin2023metric3d}, and LiDAR segmentation \cite{ zhu2021cylindrical}. To bridge the 3D representation with 2D pseudo label, they proposed specific 3D representation formats to facilitate volume rendering, including signed distance functions \cite{ huang2024selfocc} and 3D Gaussians \cite{ jiang2024gausstr}. The core strategy of these self-supervised methods is to supplement pseudo semantic supervision. They project LiDAR segmentation annotations from the same dataset into 2D pseudo labels \cite{boeder2024occflownet, pan2024renderocc}. However, this pseudo labels still is expensive and provides only sparse supervision. They also employ a 2D vision foundation model to generate 2D pseudo labels \cite{huang2024selfocc}, which suffers from class misalignment. For the overall pipeline, they supervise only the final prediction using 2D pseudo labels and do not apply separate supervision to geometric and semantic feature learning. Thus, this non-decoupled design leads to insufficient geometric learning, which reduces volume-rendering quality and weakens semantic supervision.

To address these challenges, we propose OccLE, a label-efficient 3D semantic occupancy prediction method, as illustrated in Fig. \ref{fig1}, which uses limited voxel annotations while maintaining high performance. We design the label-efficient learning paradigm: taking image and LiDAR inputs, fully decoupling semantic and geometric feature learning, fusing their feature grids for 3D semantic occupancy prediction, and supervising with limited voxel annotations and aligned pseudo labels. Specifically, OccLE consists of the semantic branch, the geometric branch, and semantic-geometric feature grid fusion. The semantic branch distills dataset-specific and open-vocabulary 2D foundation models to produce aligned pseudo labels for supervising 2D and 3D semantic learning, independently of geometry. The geometric branch integrates image and LiDAR and inputs in cross-plane synergy based on their inherent characteristics, employing semi-supervision to enhance geometry learning. The semantic-geometric feature grid fusion employs Dual Mamba to fuse the two feature grids for lightweight long-range relationship modeling, and introduces a scatter-accumulated projection to guide unannotated predictions with aligned pseudo labels, without relying on any specific feature grid format. In summary, the main contributions of this work are summarized as follows: In summary, the main contributions of this work are summarized as follows:

\noindent\textbullet\ We introduce a label-efficient 3D semantic occupancy prediction task and a learning paradigm that decouples semantic and geometric learning, followed by their synergistic integration. The model is supervised with limited voxel annotations and aligned pseudo labels to achieve high performance.

\noindent\textbullet\ We present three techniques: distilling 2D foundation model for 2D and 3D semantic learning; semi-supervised geometry learning using cross-plane image and LiDAR feature synergy; and semantic-geometric feature grids fusion with scatter-accumulated projection auxiliary supervision.

\noindent\textbullet\ Experimental results show that OccLE achieves competitive performance compared to fully supervised methods, attaining 16.59 \% and 27.53\% mIoU on the SemanticKITTI validation set and Occ3D-nuScenes using only 10 \% of the voxel annotations.

\section{Related Works}
\label{sec2}
\subsection{3D Semantic Occupancy Prediction}
\label{sec2.1}
3D semantic occupancy prediction, also referred to as the semantic scene completion (SSC), was first introduced by SSCNet \cite{song2017semantic}. In SSCNet, image features are lifted into a 3D representation to predict voxel-level semantics. With the advancement of autonomous driving \cite{li2022modality,fang2024cross}, SSC has attracted significant attention for providing clear and efficient scene representations. Based on the type of input, SSC methods can be categorized into camera-based and multi-modal methods.

For camera-based 3D semantic occupancy prediction, a key challenge lies in lifting 2D to 3D representations. Previous works \cite{zhang2023occformer,wei2023surroundocc} utilize calibration and depth information to establish explicit backwards projections. ViewFormer \cite{li2025viewformer} proposes a learning attention mechanism to aggregate multi-view features. Some studies \cite{gu2023mamba,huang2025gaussianformer} represent scenes using 3D Gaussians. TPVFormer \cite{huang2023tri} introduces three perpendicular views (TPV) to enhance the interaction of image features.

Since dimensional transformation based on estimated depth is ill-posed for camera-based methods, multi-modal method incorporates LiDAR and radar inputs to obtain accurate distance measurements. Its essential research is modal fusion. OccFusion \cite{ming2024occfusion} directly concatenates feature channels. MetaOcc \cite{yang2025metaocc} learn global and local alignment to enhance fused representations. Some studies \cite{zhang2024occfusion,zhang2025occloff} employ point-to-point queries from 3D to 2D. OccGen \cite{wang2025occgen} employs a diffusion model to refine the fused features.

Building on prior work, OccLE takes camera images and LiDAR scans as inputs, applies backward projection to derive 3D feature grids from 2D image feature maps, and proposes a lightweight fusion strategy for geometry learning, termed cross-plane image and LiDAR feature synergy.

\subsection{Label-efficient learning}
\label{sec2.2}
Label-efficient learning seeks to minimize the reliance on large amounts of label while maintaining high model performance. Prior works have proposed novel learning paradigms to achieve this, such as semi-supervised learning and self-supervised learning. Semi-supervised learning \cite{berthelot2019mixmatch} combines a small labeled dataset with abundant unlabeled data to improve model generalization, while self-supervised learning \cite{gui2024survey} generates supervision directly from unlabeled data through pretext tasks.

In label-efficient learning for 3D semantic occupancy prediction, self-supervision methods render 3D feature into 2D format, and supervise with labels from other tasks. To facilitate rendering, SelfOcc \cite{huang2024selfocc} introduces a signed distance function-based representation, while GaussTR \cite{jiang2024gausstr} proposes a 3D Gaussian-based representation. For auxiliary supervision, OccFlowNet \cite{boeder2024occflownet} utilizes projected LiDAR semantic, whereas RenderOcc \cite{pan2024renderocc} relies on estimated depth and image segmentation.

Inspired by the above, OccLE utilizes the pseudo label from image segmentation as a key supervision to enable label-efficient learning. We distill 2D foundation model for multi-modal semantic learning, while supervising the accumulated projection in the final prediction.

\section{Methodology}
\label{sec3}
\subsection{Overview}
\label{sec3.1}

\begin{figure*}[!htbp] 
\begin{center}
\includegraphics[width=\linewidth]{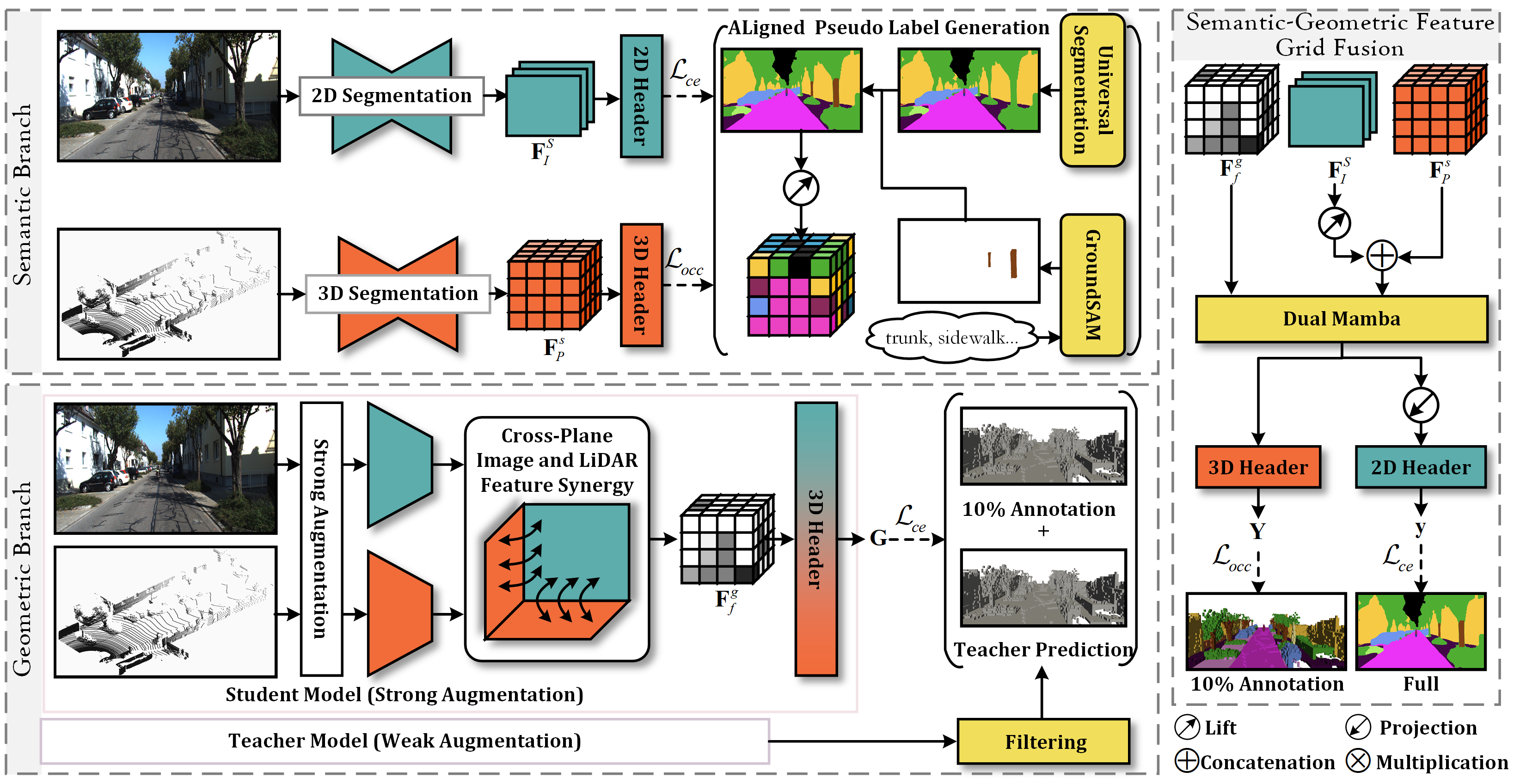}  
\end{center}
\caption{The overview of OccLE. First, we distill 2D foundation models to predict aligned pseudo labels for supervising 2D and 3D semantic learning. Next, we propose cross‐plane image and LiDAR feature synergy and apply semi‐supervision to learn geometry. Finally, we fuse semantic and geometric feature grids via Dual Mamba and supervise the unanotated prediction with aligned pseudo label using scatter‐accumulated projection.}
\label{fig2}
\end{figure*}

We aim to take multi-frame camera images ${\left\{ {{{\bf{I}}_{t - i}}} \right\}_{i = 0}^{N - 1} \in {{ \mathbb{R} }^{h \times w \times 3}}}$, multi-frame point clouds ${\left\{ {{{\bf{P}}_{t - i}}} \right\}_{i = 0}^{N - 1} \in {{ \mathbb{R} }^{n \times 4}}}$, camera calibration ${\left\{ {{\bf{K}}{\rm{,}}{\bf{T}}} \right\}}$ as inputs to predict the 3D semantic occupancy ${{{\bf{Y}}_t} \in {\left\{ {{c_0},{c_1}, \cdots ,{c_{M - 1}}} \right\}^{H \times W \times Z}}}$ at timestep ${t}$. To enable label-efficient training, we supervise the model training with limited voxel annotations (e.g., 10\% ${\left\{ {{{{\bf{\bar Y}}}_t}} \right\}}$ of all samples). Here, ${{\bf{K}}{\rm{,}}{\bf{T}}}$ represent the intrinsic and extrinsic camera matrices, respectively; ${\left\{ {{{{\bf{\bar Y}}}_t}} \right\}}$ is the ground truth of 3D semantic occupancy prediction. ${N}$ denotes frame number; ${M}$ denotes class number, ${h}$ and ${w}$ denote the height and width of camera image; ${H}$, ${W}$, and ${Z}$ represent the length, width, height of voxel grid. Our method is formulated as ${{{\bf{Y}}_t} = \Theta \left( {\left\{ {{{\bf{I}}_{t - i}}} \right\}_{i = 0}^{N - 1},\left\{ {{{\bf{P}}_{t - i}}} \right\}_{i = 0}^{N - 1},{\bf{K}}{\rm{,}}{\bf{T}}} \right)}$, where ${\Theta }$ is our proposed OccLE. For clarity, we omit the timestep subscript ${t}$ for all variables in the following description.

The overview of OccLE is illustrated in Fig. \ref{fig2}. For simplicity, only a single input frame is shown. To enable label-efficient learning, our pipeline comprises three components: a semantic branch, a geometric branch, and a semantic-geometric feature grid fusion module. We decouple the semantic and geometric learning with distinct supervision strategies, and fuse their feature grids to predict 3D semantic occupancy. In the semantic branch, we distill 2D foundation models to predict aligned pseudo labels and supervise 2D and 3D semantic feature learning. In the geometric branch, we synergize image and LiDAR features based on their inherency and employ semi‐supervision to strengthen geometry learning. In the semantic-geometric feature grid fusion module, we employ Dual Mamba to fuse feature grids and employ scatter‐accumulated projection to supervise unanotated predictions with aligned pseudo labels. 

\subsection{Distill 2D Foundation Models for 2D and 3D Semantic Learning}
\label{sec3.2}
3D semantic occupancy prediction is implemented through the entanglement of geometric and semantic features \cite{li2023voxformer}. The geometric feature indicates whether a voxel is occupied, while the semantic feature specifies the class occupying the voxel. 3D semantic occupancy prediction maps the feature grid to voxel semantic labels. Achieving high accuracy traditionally requires extensive voxel annotations to guide the mapping, but obtaining such annotations is costly. To mitigate this requirement, we notice that the image semantic segmentation maps the feature map to pixel semantic label and benefits from large annotated datasets. Therefore, we aim to distill 2D foundation model for 2D and 3D semantic learning, where the 2D foundation model predicts aligned pseudo labels to supervise the semantic learning of image and LiDAR inputs.

2D foundation models may not share the same class definitions as 3D semantic occupancy prediction. For example, a 2D foundation model trained on the Cityscapes dataset \cite{cordts2016cityscapes} does not include categories such as bicyclist or truck when applied to the SemanticKITTI dataset \cite{behley2019semantickitti}. To resolve this class unalignment, we integrate both dataset-specific and open-vocabulary 2D foundation models. The dataset-specific one produces primary labels for shared classes, while the open-vocabulary one produces auxiliary labels for unaligned classes. Specifically, we utilize MSeg \cite{lambert2020mseg} as a dataset-specific model and SAM2 \cite{ravi2024sam} for open-vocabulary model. We adapt MSeg universal class classification to the task-specific classes, input the unaligned class name prompts into SAM2, and merge outputs using Boolean operations to generate aligned pseudo labels ${{\bf{\bar s}}}$. The details of category alignment are provided in Appendix \textcolor{cvprblue}{B}.

We employ ${{\bf{\bar s}}}$ to fully supervise semantic learning of images and LiDAR, enabling lightweight deployment via distillation. Specifically, we employ a UNet \cite{ronneberger2015u} ${{{\cal G}_{s2d}}}$ to extract features map ${{\bf{F}}_I^s}$ from the image ${{\bf{I}}}$. We adopt the voxelization in VoxelNet \cite{zhou2018voxelnet} and utilize a sparse UNet3D \cite{yan2018second} ${{{\cal G}_{s3d}}}$ to extract features grid ${{\bf{F}}_P^s}$ from the LiDAR ${{\bf{P}}}$. The mathematical expression is given by:

\begin{equation}
\label{equ1}
\left\{
\begin{aligned}
\mathbf{F}_I^s &= \mathcal{G}_{s2}(\mathbf{I}) \\
\mathbf{F}_P^s &= \mathcal{G}_{s3}(\mathcal{V}(\mathbf{P}))
\end{aligned}
\right.
\end{equation}

\noindent where ${{\cal V}}$ refers to the voxelization process, which randomly samples up to 35 points per voxel. Each point includes its position, intensity, and the offset from the voxel center.

The semantic branch predicts image segmentation ${{\bf{s}}}$ and LiDAR segmentation ${{\bf{S}}}$ using a 2D header and 3D header. To supervise ${{\bf{S}}}$, we lift ${{\bf{\bar s}}}$ to ${{\bf{\bar S}}}$ as follows:

\begin{equation}
\label{equ2}
\left\{
\begin{aligned}
{{\bf{\bar S}}} &= {\cal F}\left( {{\bf{\bar s}},{\bf{p}}} \right) \\
{\bf{p}} &= {\bf{K}}\left( {{\bf{T}} \cdot {\bf{P}}} \right)
\end{aligned}
\right.
\end{equation}

\noindent where ${{\cal F}}$ denotes the sampling function, ${{\bf{P}}}$ denote the voxel grid coordinates, ${{\bf{p}}}$ represent the corresponding projected coordinates on the image plane.

\subsection{Semi-Supervised Geometry Learning}
\label{sec3.3}

Prior works \cite{li2023voxformer, wang2024h2gformer} depend on absolute depth in learning geometry for 3D semantic occupancy prediction. Because depth estimation is ill-posed, reducing voxel annotations exacerbates uncertainty in this sequential prediction. Since depth is sourced from the LiDAR projected on the image plane, we directly use image and LiDAR inputs, propose a lightweight and modality‑complementary geometry learning module, termed cross‑plane image and LiDAR feature synergy, and employ a semi‑supervision to learn an accurate geometric feature grid with limited voxel annotations.

\subsubsection{Cross-Plane Image and LiDAR Feature Synergy}
\begin{figure*}[!htbp] 
\begin{center}
\includegraphics[width=\linewidth]{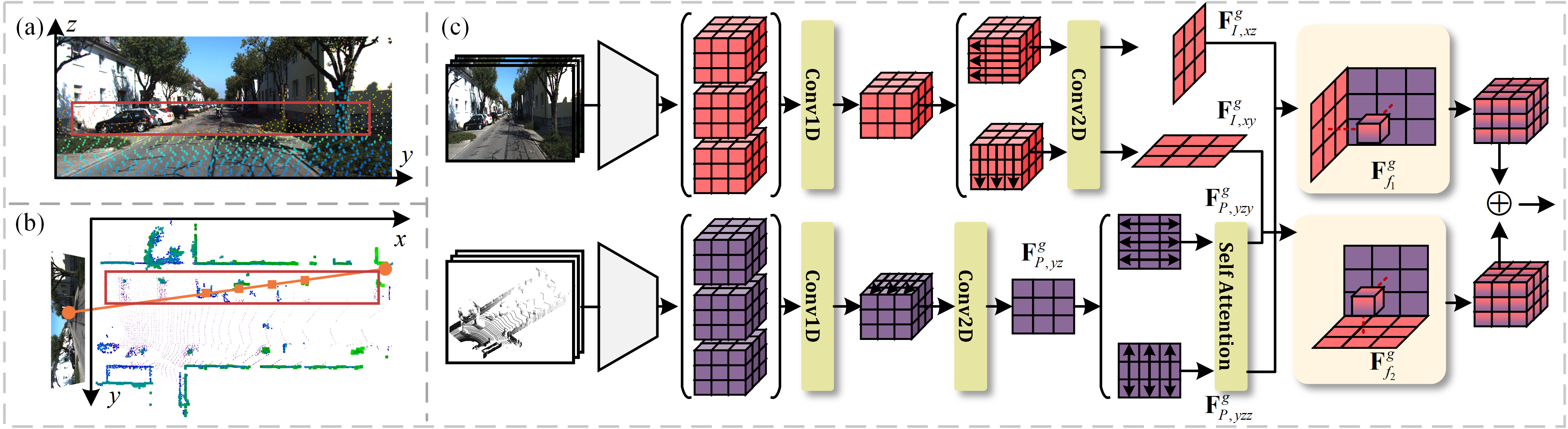}  
\end{center}
\caption{Illustration of geometry learning. (a) Frontal view feature comparison. (b) BEV view feature comparison. (c) The cross-plane image and LiDAR feature synergy.}
\label{fig3}
\end{figure*}

First, we examine the characteristics of the image and LiDAR inputs. As shown by the red boxes in Fig. \ref{fig3}\textcolor{cvprblue}{a} and Fig. \ref{fig3}\textcolor{cvprblue}{b}, the frontal-view LiDAR is sparse and omits information in the upper region, whereas the image offers continuous context. In the bird’s-eye view (BEV), the image cannot resolve distances along each ray, while the LiDAR provides accurate range measurements. In summary, the image is reliable on the ${yz}$-plane and the LiDAR is reliable along the ${x}$-axis in world coordinates. Motivated by their inherency and inspired by \cite{zhang2024lightweight}, we propose using TPV to achieve a synergistic and lightweight multi-modal features fusion by projecting features onto their optimal TPV planes and applying multiplicative integration.

The cross-plane image and LiDAR feature synergy is illustrated in Fig. \ref{fig3}\textcolor{cvprblue}{c}. We voxelize multi-frame LiDAR using the VoxelNet \cite{zhou2018voxelnet} and process them with a sparse 3D encoder to obtain LiDAR feature grids. In parallel, we extract feature maps from multi-frame images using a 2D encoder and lift into the feature grid defined in Equ. \ref{equ2}. We fuse the multi-frame feature grids into a single grid via a Conv1D layer. For the LiDAR feature grid, we stack along the ${y}$ and ${z}$ axes and apply a Conv2D layer to produce feature maps ${{\bf{F}}_{I,xz}^g \in {{ \mathbb{R} }^{{H_1} \times {Z_1}}}}$ and ${{\bf{F}}_{I,xy}^g \in {{ \mathbb{R} }^{{W_1} \times {Z_1}}}}$. For the image feature grid, we stack along the ${x}$-axis and apply a Conv2D layer to generate feature map ${{\bf{F}}_{P,yz}^g \in {{ \mathbb{R} }^{{H_1} \times {W_1}}}}$. To enhance ${{\bf{F}}_{P,xy}^g \in {{ \mathbb{R} }^{{H_1} \times {W_1}}}}$, we apply self-attention ($\texttt{SA}$) along the ${y}$ and ${z}$ axes, yielding feature maps ${{\bf{F}}_{P,yzy}^g \in {{ \mathbb{R} }^{{H_1} \times {W_1}}}}$, ${{\bf{F}}_{P,yzz}^g \in {{ \mathbb{R} }^{{H_1} \times {W_1}}}}$ as follows:

\begin{equation}
\label{equ3}
\small
\left\{ \begin{array}{l}
{\bf{F}}_{P,yzy}^g = {\bf{F}}_{P,yz}^g + SA\left( {{\bf{F}}_{P,yz}^g} \right)\\
{\bf{F}}_{P,yzz}^g = {\bf{F}}_{P,yz}^g + SA{\left( {{\bf{F}}{{_{P,yz}^g}^T}} \right)^T}
\end{array} \right.
\end{equation}

We fuse the image and LiDAR feature maps via matrix multiplication in the TPV. The fused feature grids ${{\bf{F}}_{{f_1}}^g,{\bf{F}}_{{f_2}}^g \in {{ \mathbb{R} }^{{H_1} \times {W_1} \times {Z_1}}}}$ are defined as follows:

\begin{equation}
\label{equ4}
\left\{ \begin{array}{l}
{\bf{F}}_{{f_1}}^g = {\bf{F}}_{I,xz}^g \times {\bf{F}}_{P,yzz}^g\\
{\bf{F}}_{{f_2}}^g = {\bf{F}}_{I,xy}^g \times {\bf{F}}_{P,yzy}^g
\end{array} \right.
\end{equation}

We add ${{\bf{F}}_{{f_1}}^g}$ and ${{\bf{F}}_{{f_2}}^g}$ to obtain the geometric feature grid ${{\bf{F}}_{{f}}^g}$ and predict geometry ${{\bf{G}} \in {{ \mathbb{R} }^{H \times W \times Z}}}$ via a 3D header.

In the proposed workflow, feature grids are converted into feature maps by stacking and subsequently reconstructed into feature grids during the final fusion. In contrast to the 3D attention mechanisms adopted in \cite{li2023voxformer, wang2024h2gformer, xue2024bi}, the cross-plane image and LiDAR feature synergy employs only Conv1D, Conv2D, and \texttt{SA} (with computational complexity ${{\cal O}\left( {{H_1}^2} \right)}$ or ${{\cal O}\left( {{W_1}^2} \right)}$) to learn features, making it more lightweight and suitable for real-time applications.

\subsubsection{Semi-Supervised Pipeline}
To facilitate cross-plane image and LiDAR feature synergy training with limited voxel annotations, we adopt the weak-to-strong consistency principle \cite{yang2023revisiting} within a semi-supervised learning. The teacher and student models share the same structure. Table \ref{table1} lists their data-augmentation strategy.

\begin{table}[!htb]
  \centering
  \footnotesize
  
  \caption{Data augmentation strategy. The number of signs indicates the augmentation strength.}
  \label{table1}
  
  \setlength{\tabcolsep}{2pt}  
  \resizebox{1.0\linewidth}{!}{%
  \begin{tabular}{l|lllll|ll}
    \toprule
    \multicolumn{1}{l|}{} & \multicolumn{5}{c|}{Image} & \multicolumn{2}{c}{LiDAR} \\
    \cmidrule(lr){2-6} \cmidrule(lr){7-8}
    & Brightness & HSV space & Motion blur & Weather sim. & Cutout & Point dropout & Voxelization \\
    \midrule
    \multirow{1}{*}{Teacher}  & \checkmark & \checkmark & \ding{55} & \ding{55} & \ding{55} & \checkmark & \checkmark \\
    \midrule
    \multirow{1}{*}{Student} & \checkmark \checkmark \checkmark & \checkmark \checkmark \checkmark& \checkmark \checkmark & \checkmark \checkmark & \checkmark & \checkmark \checkmark \checkmark & \checkmark \checkmark \checkmark\\
    \bottomrule
  \end{tabular}
}
\end{table}

We classify image augmentations by their effects on features. Brightness, contrast, and HSV adjustments modify global characteristics and constitute weak augmentation. Weather simulation and motion blur degrade feature clarity and serve as middle augmentation. Cutout removes entire regions of the feature map and thus represents strong augmentation. For LiDAR, we vary augmentation strength by adjusting the dropout probability, the dropout number, and the sampling number.

During knowledge distillation, we filter the predictions of teacher model based on confidence scores, which are computed as the maximum channel prediction for each sample. We then concatenate the filtered predictions with the limited voxel annotations to supervise the student model.

\subsection{Semantic-Geometric Feature Grid Fusion}
\label{sec3.4}

In this section, we fuse semantic and geometric feature grids to predict 3D semantic occupancy. Previous works \cite{wang2024h2gformer,xue2024bi} employs attention mechanisms to enable long-range interactions within the feature grid. However, with sequence length ${HWZ}$, 3D ${\texttt{DA}}$ still incurs substantial computational overhead. Recent work  \cite{li2024occmamba} adapts Mamba \cite{gu2023mamba} module for efficient long-range dependency learning in this task. Inspired by this approach and given the spatial alignment of our semantic and geometric feature grids, we propose a Dual Mamba that process semantic and geometric feature grids independently, exchanges subsets of their channels for fusion. Specifically, we lift the feature map ${{\bf{F}}_I^s}$ to a feature grid via Equ. \ref{equ2} and combine it with ${{\bf{F}}_P^s}$ to yield the semantic feature grid ${{\bf{F}}_f^s}$, Dual mamba then takes ${{\bf{F}}_f^s}$ and ${{\bf{F}}_f^g}$ as input and outputs the fused feature grid ${{\bf{F}}_{3d}^f}$. The detailed structure of Dual Mamba is illustrated in Appendix \textcolor{cvprblue}{A}.

GaussTR \cite{jiang2024gausstr} projects Gaussian feature grids onto the image plane and supervises predictions using pseudo labels from 2D segmentation. Inspired by this approach and aiming to leverage the aligned pseudo label ${{\bf{\bar s}}}$ for supervising 3D semantic occupancy prediction, we propose a simple and effective method, termed scatter-accumulated projection, as follows:

\begin{equation}
\label{equ6}
{\bf{F}}_{2d}^f = \sum\limits_{\bf{p}} {{\bf{F}}_{3d}^f} 
\end{equation}

\noindent where ${\sum\limits_{\bf{p}} {\left(  \cdot  \right)} }$ denotes scattered accumulation over index ${{\bf{p}}}$. This operation projects ${{\bf{F}}_{3d}^f}$ onto the image plane and aggregates projected feature along each ray. It enables the model to identify the semantic class of each feature grid and to capture occlusion and occupancy states in geometry, thereby facilitating both semantic and geometric supervision. Additionally, this method is agnostic to the specific format of the feature grid, thus offering better scalability.

We use a 3D and 2D header to predict the 3D semantic occupancy ${{\bf{Y}}}$ and the projected segmentation ${{\bf{y}} \in {{ \mathbb{R} }^{h \times w}}}$ from feature grid ${{\bf{F}}_{3d}^f}$ and projected feature map ${{\bf{F}}_{2d}^f}$.

\subsection{Losses}
\label{sec3.5}

During training, we define the loss functions for the semantic branch ${{\ell _{sem}}}$, geometric branch ${{\ell _{geo}}}$, and semantic-geometric feature grid fusion ${{\ell _{fus}}}$ as follows:

\begin{equation}
\label{equ5}
\left\{ \begin{array}{l}
{\ell _{sem}} = {\ell _{ce}}\left( {{\bf{s}},{\bf{\bar s}}} \right) + {\ell _{occ}}\left( {{\bf{S}},{\bf{\bar S}}} \right)\\
{\ell _{geo}} = {\ell _{ce}}\left( {{\bf{G}},{\bf{\bar G}}} \right)\\
{\ell _{fus}} = {\ell _{occ}}\left( {{\bf{Y}},{\bf{\bar Y}}} \right) + {\ell _{ce}}\left( {{\bf{y}},{\bf{\bar s}}} \right)
\end{array} \right.
\end{equation}

\noindent where ${{\ell _{ce}}}$ denotes the weighted cross-entropy loss, with weights computed from class frequencies. ${{\ell _{occ}}}$ is a comprehensive occupancy loss, following prior work \cite{li2023voxformer, jiang2024symphonize}. ${{\bf{\bar G}}}$ represents the ground-truth of geometry. We apply ${{\ell _{occ}}\left( {{\bf{S}},{\bf{\bar S}}} \right)}$ only to voxels with more than one point to supervise voxel-level semantics.

\begin{table*}[!htb]
\centering
\caption{Quantitative results on SemanticKITTI validation set. \textbf{Bold} and \underline{underline} represent the best and second best results, respectively. Inp. and Sup. indicate the input modality and the supervision type, respectively. C and L denote the camera and LiDAR inputs, respectively.}
\label{table2}
\setlength{\tabcolsep}{3pt} 
\resizebox{1.0\textwidth}{!}{%
\begin{tabular}{l|l|l|l>{\columncolor{gray!30}}l|lllllllllllllllllll}
\toprule
Method & Inp. &Sup. & IoU & mIoU &\rotatebox{90}{{\tikz \fill[c1] (0,0) rectangle (0.6em,0.6em);} car (3.92\%)} & \rotatebox{90}{{\tikz \fill[c2] (0,0) rectangle (0.6em,0.6em);} bicycle (0.03\%)} & \rotatebox{90}{{\tikz \fill[c3] (0,0) rectangle (0.6em,0.6em);} motorcycle (0.03\%)} & \rotatebox{90}{{\tikz \fill[c4] (0,0) rectangle (0.6em,0.6em);} truck (0.16\%)} & \rotatebox{90}{{\tikz \fill[c5] (0,0) rectangle (0.6em,0.6em);} other-veh. (0.20\%)} & \rotatebox{90}{{\tikz \fill[c6] (0,0) rectangle (0.6em,0.6em);} person (0.07\%)} & \rotatebox{90}{{\tikz \fill[c7] (0,0) rectangle (0.6em,0.6em);} bicyclist (0.07\%)} & \rotatebox{90}{{\tikz \fill[c8] (0,0) rectangle (0.6em,0.6em);} motorcyclist (0.05\%)} & \rotatebox{90}{{\tikz \fill[c9] (0,0) rectangle (0.6em,0.6em);} road (15.30\%)} & \rotatebox{90}{{\tikz \fill[c10] (0,0) rectangle (0.6em,0.6em);} parking (1.12\%)} & \rotatebox{90}{{\tikz \fill[c11] (0,0) rectangle (0.6em,0.6em);} sidewalk (11.13\%)} & \rotatebox{90}{{\tikz \fill[c12] (0,0) rectangle (0.6em,0.6em);} other-grnd.(0.56\%)} & \rotatebox{90}{{\tikz \fill[c13] (0,0) rectangle (0.6em,0.6em);} building (14.10\%)} & \rotatebox{90}{{\tikz \fill[c14] (0,0) rectangle (0.6em,0.6em);} fence (3.90\%)} & \rotatebox{90}{{\tikz \fill[c15] (0,0) rectangle (0.6em,0.6em);} vegetation (39.3\%)} & \rotatebox{90}{{\tikz \fill[c16] (0,0) rectangle (0.6em,0.6em);} trunk (0.51\%)} & \rotatebox{90}{{\tikz \fill[c17] (0,0) rectangle (0.6em,0.6em);} terrain (9.17\%)} & \rotatebox{90}{{\tikz \fill[c18] (0,0) rectangle (0.6em,0.6em);} pole (0.29\%)} & \rotatebox{90}{{\tikz \fill[c19] (0,0) rectangle (0.6em,0.6em);} traf.-sign (0.08\%)} \\ 
\midrule
TPVFormer \cite{huang2023tri} & C & Full & 35.61 & 11.36 & 23.81 & 0.36 & 0.05 & 8.08 & 4.35 & 0.51 & 0.89 & 0.00 & 56.50 & 20.60 & 25.87 & 0.85 & 13.88 & 5.94 & 16.92 & 2.26 & 30.38 & 3.14 & 1.52 \\
OccFormer \cite{zhang2023occformer}& C & Full& 36.50 & 13.46 & 25.09 & 0.81 & 1.19 & \textbf{25.53} & 8.52 & 2.78 &2.82 & 0.00 & 58.85 & 19.61 & 26.88 & 0.31 & 14.40 & 5.61 & 19.63 & 3.93 & 32.62 & 4.26 & 2.86 \\
VoxFormer \cite{li2023voxformer} & C & Full& 44.15 & 13.35 & 25.64 & 1.28 & 0.56 & 7.26 & 7.81 & 1.93 & 1.97 & 0.00 & 53.57 & 19.69 & 26.52 & 0.42 & 19.54 & 7.31 & 26.10 & 6.10 & 33.06 & 9.15 & 4.94 \\
PanoSSC \cite{shi2024panossc}& C& Full & 34.94 & 11.22 & 19.63 & 0.63 & 0.36 & 14.79 & 6.22 & 0.87 & 0.00 & 0.00 & 56.36 & 17.76 & 26.40 & \underline{0.88} & 14.26 & 5.72 & 16.69 & 1.83 & 28.05 & 1.94 & 0.70 \\
H2GFormer \cite{wang2024h2gformer}& C & Full& 44.69 & 14.29 & 28.21 & 0.95 & 0.91 & 6.80 & 9.32 & 1.15 & 0.10 & 0.00 & 57.00 & \underline{21.74} & 29.37 & 0.34 & 20.51 & 7.98 & 27.44 & 7.80 & 36.26 & 9.88 & 5.81 \\
OctreeOcc \cite{lu2023octreeocc}& C & Full& 44.71 & 13.12 & 28.07 & 0.64 & 0.71 & 16.43 & 6.03 & 2.25 & 2.57 & 0.00 & 55.13 & 18.68 & 26.74 & 0.65 & 18.69 & 4.01 & 25.26 & 4.89 & 32.47 & 3.72 & 2.36 \\
SGN \cite{mei2024camera}& C & Full& \underline{46.21} & 15.32 & 33.31 & 0.61 & 0.46 & 6.03 & 9.84 & 0.47 & 0.10 & 0.00 & 59.10 & 19.05 & 29.41 & 0.33 & 25.17 & 9.96 & 28.93 & 9.58 & 38.12 & 13.25 & 7.32 \\
Symphonies \cite{jiang2024symphonize}& C & Full & 41.44 & 13.44 & 27.23 &1.44 & 2.28 & 15.99 &9.52 & 3.19 & \textbf{8.09} & 0.00 & 55.78 & 14.57 & 26.77 & 0.19 & 18.76 & 6.18 & 24.50 & 4.32 & 28.49 & 8.99 & 5.39 \\
HASSC \cite{wang2024not}& C & Full& 44.55 & 15.88& 30.64 & 1.20 & 0.91 & \underline{23.72} & 7.77 & 1.79 & 2.47 & 0.00 & 62.75 & 20.20 & 32.40 & 0.51 & 22.90 & 8.67 & 26.47 & 7.14 & 38.10 & 9.00 & 5.23 \\  
RenderOcc \cite{pan2024renderocc}& C & Full &- & 12.87  & 24.90 & 0.37 & 0.28 & 6.03 & 3.66 & 1.91 & 3.11 & 0.00 & 57.2 & 16.11 & 28.44 & \textbf{0.91} & 18.18 & 9.10 & 26.23 & 4.87 & 33.61 & 6.24 & 3.38 \\ 
OccLoff \cite{zhang2025occloff} &C+L &Full & - &\textbf{22.62} & \textbf{46.44} &2.08 &\underline{3.91} &20.38 &8.72 &\textbf{3.88} &4.35 &0.00 &\textbf{66.25} &21.07 &\textbf{43.51} &0.57 &\textbf{41.23} &\textbf{15.86} &\textbf{41.20} &\textbf{20.06} &\textbf{46.21} &\textbf{27.60} &\textbf{16.47} \\
OccFusion \cite{ming2024occfusion} & C+L & Full &\textbf{58.68} &\underline{21.92}  &\underline{45.62} & \textbf{2.96} & 3.51 & 20.05 & 8.76 & 3.16 & \underline{4.37} & 0.00 & \underline{65.67} & \textbf{23.08} & \underline{36.33} & 0.00 & \underline{39.09} & \underline{15.70} & \underline{40.68} & \underline{19.37} & \underline{45.53} & \underline{27.57} & \underline{15.21} \\
OccGen \cite{wang2025occgen} & C+L & Full& 36.87 & 13.74  & 26.83 & 1.60 & 2.53 & 15.49 & \textbf{12.83} & \underline{3.20} &3.37 &0.00 &61.28 &20.42 &28.30 &0.43 &14.49 &6.94 &20.04 &3.94 &32.44 &4.11 &2.77 \\
OccFusion \cite{ming2024occfusion} & C+L & 10\%  &30.36 &6.03 &11.24 & 0.08 &0.09 &0.64 & 0.54 & 0.53 & 0.09 & 0.00 & 39.18 &3.23 & 15.91 & 0.06 & 8.42 & 1.09 &13.51 & 0.28 & 18.30 & 1.28 &0.07 \\
RenderOcc \cite{pan2024renderocc}& C & Self &- & 8.24  & 14.83 & 0.42 & 0.17 & 2.47 & 1.78 & 0.94 & 3.20 & 0.00 & 43.64 & 12.54 & 19.10 & 0.00 & 11.59 & 4.71 & 17.61 & 1.48 & 20.01 & 1.17 & 0.88 \\
\midrule
OccLE(Ours)& C+L & 10\% & 40.60 & 16.59 & 37.27 & \underline{2.76} & \textbf{4.39} & 0.90 & \underline{10.88} & 2.52 & 0.09 & 0.00 & 55.79 & 19.96& 28.85& 0.14 & 24.56& 12.53 & 35.19 & 16.98 & 35.29& 17.16 & 10.01\\ 
 
\bottomrule
\end{tabular}
}
\end{table*}

\begin{table*}[!htb]
\centering
\caption{Quantitative results on Occ3D-nuScenes validation set. ${\dagger}$ indicates the use of camera-visible masks during training.}
\label{table3}
\setlength{\tabcolsep}{1mm} 
\resizebox{\linewidth}{!}{%
\begin{tabular}{l|l|l|>{\columncolor{gray!30}}l|lllllllllllllllllll}
\toprule
Method & Inp. & Sup.  & mIoU
& \rotatebox{90}{\tikz{\fill[othercol] (0,0) rectangle (0.6em,0.6em);} other}
& \rotatebox{90}{\tikz{\fill[barriercol] (0,0) rectangle (0.6em,0.6em);} barrier}
& \rotatebox{90}{\tikz{\fill[bicyclecol] (0,0) rectangle (0.6em,0.6em);} bicycle}
& \rotatebox{90}{\tikz{\fill[buscol] (0,0) rectangle (0.6em,0.6em);} bus}
& \rotatebox{90}{\tikz{\fill[carcol] (0,0) rectangle (0.6em,0.6em);} car}
& \rotatebox{90}{\tikz{\fill[convcol] (0,0) rectangle (0.6em,0.6em);} cons.\ veh.}
& \rotatebox{90}{\tikz{\fill[motorcyclecol] (0,0) rectangle (0.6em,0.6em);} motorcycle}
& \rotatebox{90}{\tikz{\fill[pedestriancol] (0,0) rectangle (0.6em,0.6em);} pedestrian}
& \rotatebox{90}{\tikz{\fill[trafficconecol] (0,0) rectangle (0.6em,0.6em);} traffic cone}
& \rotatebox{90}{\tikz{\fill[trailercol] (0,0) rectangle (0.6em,0.6em);} trailer}
& \rotatebox{90}{\tikz{\fill[truckcol] (0,0) rectangle (0.6em,0.6em);} truck}
& \rotatebox{90}{\tikz{\fill[drivecol] (0,0) rectangle (0.6em,0.6em);} drive.\ surf.}
& \rotatebox{90}{\tikz{\fill[otherflatcol] (0,0) rectangle (0.6em,0.6em);} other flat}
& \rotatebox{90}{\tikz{\fill[sidewalkcol] (0,0) rectangle (0.6em,0.6em);} sidewalk}
& \rotatebox{90}{\tikz{\fill[terraincol] (0,0) rectangle (0.6em,0.6em);} terrain}
& \rotatebox{90}{\tikz{\fill[manmadecol] (0,0) rectangle (0.6em,0.6em);} manmade}
& \rotatebox{90}{\tikz{\fill[vegetationcol] (0,0) rectangle (0.6em,0.6em);} vegetation} \\\midrule

MonoScene \cite{cao2022monoscene}           &C   & Full     & 6.06 &1.75  & 7.23  & 4.26   & 4.26   & 4.93   & 9.38  & 3.98   & 3.98  & 3.90  & 4.45  & 7.17  & 14.91 & 6.32  & 7.92  & 7.43   &  1.01 & 7.43 \\
OccFormer \cite{zhang2023occformer}         &C   & Full     &21.93 &5.95  & 21.93 &5.85    & 37.83  & 17.87  & 40.44 & 42.43  & 7.36  & 23.88 & 21.81 & 20.98 & 22.38 & 30.70 & 55.35 & 28.36  & 30.70 & 18.0  \\
BEVFormer \cite{li2024bevformer}            &C   & Full     &26.88 &5.94  & 30.29 & 12.32  & 34.40  & 39.17  & 14.44 & 16.45  & 17.22 & 9.27  & 13.90 & 26.36 & 50.99 & 30.96 & 34.66 & 22.73  & 6.76  & 6.97 \\
TPVFormer \cite{huang2023tri}               &C   & Full     &27.83 &7.22  & 38.90 & 13.67  & 40.78  & 45.90  & 17.23 & 19.99  & 18.85 & 14.30 & 26.69 & 34.17 & 55.65 & 35.47 & 37.55 & 30.70  & 19.40 & 16.78 \\
FB-OCC${\dagger}$ \cite{li2023fb}                     &C   & Full     &42.06 &7.22  & 38.90 & 13.67  & 40.78  & 45.90  & 17.23 & 19.99  & 18.85 & 14.30 & 26.69 & 34.17 & 55.65 & 35.47 & 37.55 & 30.70  & 19.40 & 16.78 \\    
OccLoff${\dagger}$ \cite{zhang2025occloff}            &C+L & Full     &49.36 &\underline{13.26} &53.72  &\underline{33.20}   &55.21   & 58.94  &34.26  & \underline{43.13}  & 49.28 & 35.61 & 41.44 &48.78  &\underline{83.72}  &44.68  &\underline{57.33}  &\underline{60.15}   &63.89  &62.45 \\
SDGOcc${\dagger}$ \cite{duan2025sdgocc}               &C+L & Full     &\underline{51.66} &13.21 &\underline{57.77}  & 24.30  &\underline{60.33 } & \underline{64.28}  & \textbf{36.21} & 39.44  & \underline{52.36} &\underline{35.80}& \textbf{50.91} & \underline{53.65} & \textbf{84.56} & \textbf{47.45} & \textbf{58.00} & \textbf{61.61}  & \textbf{70.67} & 67.65 \\ 
GaussianFormer3D${\dagger}$ \cite{zhao2025gaussianformer3d} &C+L&Full &46.40 &9.80  &50.00  & 31.30  & 54.00  & 59.40  & 28.10 & 36.20  &46.20  & 26.70 & 40.20 & 49.70 & 79.10 & 37.30 & 49.00 & 55.00  &69.10  &67.60 \\
OccFusion${\dagger}$ \cite{ming2024occfusion}         &C+L & Full     &46.79 &11.65 & 47.81 & 32.07  & 57.27  &57.51   &31.80  & 40.11  & 47.35 & 33.74 &45.81 & 50.35 & 78.79 & 37.17 & 44.36 &53.36   &63.18  &63.20 \\
DAOcc${\dagger}$\cite{yang2024daocc}                  &C+L & Full     &\textbf{53.82} &12.40 &\textbf{59.60}  &\textbf{38.40}   &\textbf{61.90}  & \textbf{67.10}  &\underline{35.30}  & \textbf{48.20}  & \textbf{59.10} & \textbf{43.50} &\underline{50.90}  &\textbf{56.30}  &83.00  &\underline{44.70}  &56.70  &59.90   &\underline{70.00} &\underline{68.10} \\
SelfOcc \cite{huang2024selfocc}             &C   & Self &9.30  &0.00  & 0.15  & 0.66   & 5.46   & 12.54  & 0.00  & 0.80   & 2.10  & 0.00  & 0.00  & 8.25  & 55.49 &0.00   & 26.30 & 26.54 & 14.22 & 5.60 \\
OccNeRF \cite{zhang2023occnerf}             &C   & Self &9.53  &0.00  & 0.83  & 0.82   & 5.13   & 12.49  & 3.50  & 0.23   & 3.10  & 1.84  & 0.52  & 3.90  & 52.62 &0.00   & 20.81 & 24.75 & 18.45 & 13.19 \\
DistillNeRF \cite{wang2024distillnerf}      &C   & Self &8.93  &0.03  & 1.35  & 2.08   & 10.21  & 10.09  & 2.56  & 1.98   & 5.54  & 4.62  & 1.43  & 7.90  & 43.02 &0.00   & 16.86 & 15.02 & 14.06 & 15.06 \\
GaussTR \cite{jiang2024gausstr}             &C   & Self&11.70 &-     & 2.09  & 5.22   & 14.07  & 20.43  & 5.70  & 7.08   & 5.12  & 3.93  & 0.92  & 13.36 & 39.44 &-      & 15.68 & 22.89 & 21.17 & 21.87 \\
RenderOcc \cite{pan2024renderocc}           &C   & Self &23.93 &5.69  & 27.56 & 14.36  & 19.91  & 20.56  & 11.96 & 12.42  & 12.14 & 14.34 & 20.81 & 18.94 & 68.85 &33.35  & 42.01 & 43.94 & 17.36 &22.61 \\ \midrule
OccLE (Ours)                                &C+L & 10\%    &27.53 &\textbf{28.58}  & 0.00 & 27.33  & 37.44  & 8.13  & 0.00 & 17.76  & 0.00 &12.56 & 20.18 & 71.51 & 30.97 &38.46  & 38.94& 27.90 & 24.37 &\textbf{83.95} \\ \bottomrule
\end{tabular}
}
\end{table*}

\section{Experiment}
\label{sec4}
\subsection{Dataset and Metric}
\label{sec4.1}
\subsubsection{Dataset}
We evaluate OccLE on the SemanticKITTI \cite{behley2019semantickitti} and Occ3D‑nuScenes \cite{tian2023occ3d}. SemanticKITTI annotates scenes of size ${51.2m \times 51.2m \times 6.4m}$ with ${0.2 m}$ voxel size for 20 classes (19 semantics + 1 free). Occ3D-nuScenes annotate the ${80.0m \times 80.0m \times 6.4m}$ with ${0.4m}$ voxel size for 18 classes (17 semantics + 1 free). To simulate a label-efficient learning, we use only 10\% of the voxel annotations in the training set, obtained via interval sampling. 
\subsubsection{Metric}
There are two domain metrics: intersection over union (IoU) and mean intersection over union (mIoU) across 19 classes. IoU measures the overall scene completion quality, while mIoU evaluates the quality of semantic segmentation for each class. We use IoU as the primary metric for the geometric branch and mIoU as the primary metric for the semantic branch and semantic-geometric feature grid fusion.

\subsection{Implementation Details}
\label{sec4.2}
All experiments are conducted on 2 NVIDIA A6000 GPUs with a batch size of 1. The AdamW optimizer is used with an initial learning rate of ${2{e^{ - 4}}}$ and a weight decay of ${{e^{ - 4}}}$, All training durations are set to 40 epochs, while the teacher model of the geometric branch is set to 100 epochs. More details are presented in Appendix \textcolor{cvprblue}{D.2}.

\subsection{Main Result}
\label{sec4.3}

We present a quantitative comparison on the SemanticKITTI validation set and Occ3D‑nuScenes validation set in Table \ref{table2} and Table \ref{table3}, respectively. For the SemanticKITTI dataset, OccLE obtains competitive performance, achieving 16.59 \% mIoU and 39.96 \% IoU. It surpasses all fully supervised camera-based methods (HASSC \cite{wang2024not} with 40.66 \% mIoU) and approaches the performance of fully supervised multi-modal methods (OccLoff \cite{zhang2025occloff} with 22.62 \% mIoU). For the Occ3D‑nuScenes dataset, OccLE is trained using camera-visible masks for 10\% of the training samples and a single-frame input. Please note that using camera-visible masks can increase other methods from ~20 to ~40 mIoU. Despite this noted gap, the experimental results show that OccLE achieves 27.53 mIoU, outperforming all self-supervised methods (RenderOcc \cite{pan2024renderocc} with 23.93\% mIoU) and approaching the performance of fully supervised methods. The performance comparison with other voxel annotation ratio are provided in Sec. \ref{sec4.4.4}.

\begin{figure*}[!htbp] 
\begin{center}
\includegraphics[width=\linewidth]{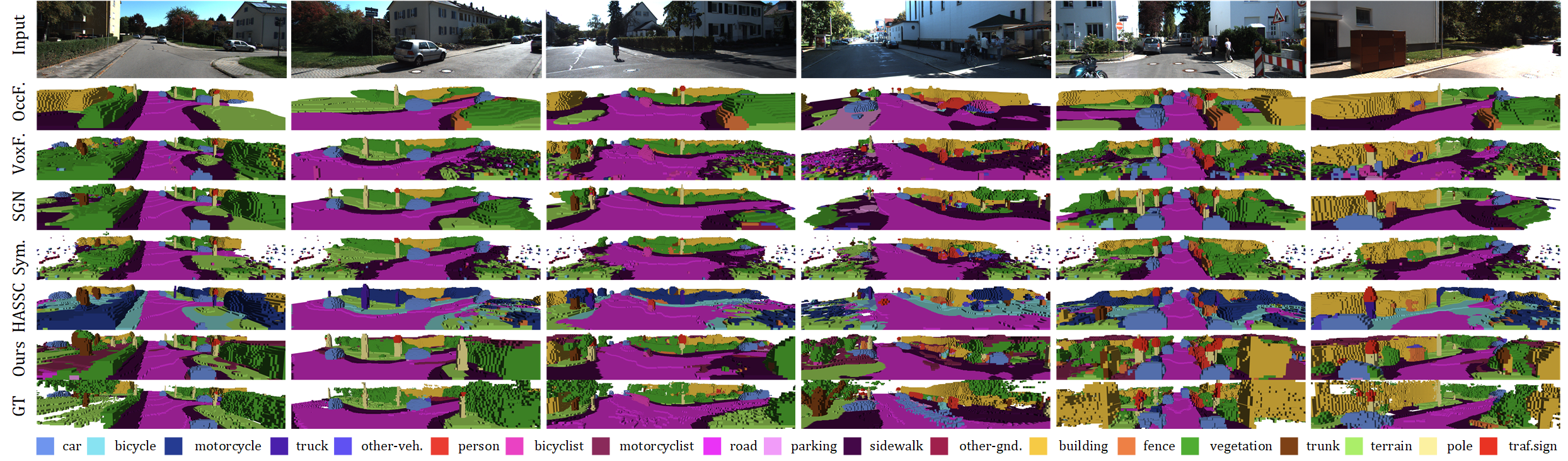}  
\end{center}
\caption{Qualitative results on the SemanticKITTI validation set. OccF., VoxF., SGN, Sym., and HASSC represent the prediction results from \cite{zhang2023occformer}, \cite{li2023voxformer}, \cite{mei2024camera}, \cite{jiang2024symphonize}, and \cite{wang2024not}, respectively. GT denotes the ground truth.}
\label{fig4}
\end{figure*}

As illustrated in Figure \ref{fig4}, OccLE achieves strong performance in complex scenes, accurately reconstructing buildings and vegetation, and distinguishing small objects such as traffic signs and poles. Additional comparisons on the SemanticKITTI hidden test set and SSCBenchKITTI-360 \cite{li2024sscbench} validation set are presented in Appendix \textcolor{cvprblue}{D.3}. Additional inference time comparisons are provided in the Appendix \textcolor{cvprblue}{D.4} to demonstrate the efficiency advantage of our method.

\subsection{Ablation Studies}
\label{sec4.4}
We conduct all ablation studies on the validation set of the SemanticKITTI dataset. Additional ablation studies are presented in the Appendix \textcolor{cvprblue}{D.5}. 

\begin{table}[H]
\centering
\setlength{\tabcolsep}{1mm} 
\caption{Ablation study of the semantic branch. The unit of Inf. Time is in milliseconds.}
\label{table4}
\resizebox{\linewidth}{!}{
\setlength{\tabcolsep}{2pt}
\begin{tabular}{ll>{\columncolor{gray!30}}lll>{\columncolor{gray!30}}l}
\toprule
${{{\cal G}_{s2d}}}$ & Inf. Time & mIoU(${{\bf{s}}}$)        & ${{{\cal G}_{s3d}}}$     & Inf. Time           & mIoU(${{\bf{S}}}$) \\ \midrule
ESPNetv2          & \textbf{4.38}             & 28.71                     & 1-layer                  &  \textbf{41.77}               & 12.38 \\
MobileNet         & 6.10             & 48.76                     & 2-layer                  &  43.22               & 13.30 \\
Ours (w/o align.) & 6.87             & 41.89                     & Ours (w/o align.)        &  44.59               & 9.89  \\
Ours (w/ align.)  & 6.87             & \textbf{51.51}             & Ours (w/ align.)         &  44.59                & \textbf{13.94} \\ \bottomrule
\end{tabular}
}
\end{table}

\subsubsection{The Semantic Branch}
\label{sec4.4.1}
To evaluate the extractor and pseudo label in the semantic branch, for comparison, we adopt ESPNetv2 \cite{mehta2019espnetv2} and MobileNet \cite{howard2017mobilenets} as ${{{\cal G}_{s2d}}}$, employ 1-layer and 2-layer versions of sparse UNet3D as ${{{\cal G}_{s3d}}}$, and train using unaligned pseudo labels. As depicted in Table \ref{table4}, 2D extractors designed for edge device exhibit a clear drop in performance, whereas our method achieves the highest mIoU(${{\bf{s}}}$) while remaining lightweight. Deeper sparse UNet3D architectures improves mIoU(${{\bf{S}}}$) without substantially increasing computational cost. Aligned pseudo labels significantly improve mIoU by offering supervision for unaligned classes. We also evaluate the effect of the semantic alignment design and semantic noise on the overall OccLE performance in Appendix \textcolor{cvprblue}{D.5 Part 1}.

\subsubsection{The Geometric Branch}
\label{sec4.4.2}
To evaluate the effects of model architecture and training strategy in the geometric branch, we design a baseline that employs Conv3D and 3D \texttt{DA} for feature extraction and interaction, and compare our semi-supervision against full supervision. As depicted in Table \ref{table5}, our cross-plane image and LiDAR feature synergy only has a minor precision drop while significantly reducing module inference time from 138.62 ms to 26.08 ms. Our data augmentation strategy boosts student model performance as augmentation degree increases. Under semi-supervision, our student model reaches 53.60 IoU. We also evaluate the case of modality failure to demonstrate the superiority of image–LiDAR feature synergy in Appendix \textcolor{cvprblue}{D.5 Part 2}.

\subsubsection{Overall Pipeline}
\label{sec4.4.3}
We use only 10\% voxel annotations for supervision as the baseline. As depicted in Table \ref{table6}, our method increases mIoU from 9.64 to 16.59 by incorporating a decoupled design and 2D-aligned pseudo labels, closely approaching the fully supervised performance of 20.32 mIoU. Using only aligned pseudo labels significantly reduces precision, indicating pattern collapse when voxel annotations are missing. Furthermore, ablating any module in OccLE results in a precision drop, demonstrating the essential role of these three modules in label-efficient learning.

\begin{table}[H]
\centering                 
\caption{Ablation study of the geometric branch, where “weak,” “medium,” and “strong” denote the degrees of data augmentation, -S and -T represents the student and teacher models.}
\label{table5}
\setlength{\tabcolsep}{1mm} 
\begin{tabular}{llllll>{\columncolor{gray!30}}l}
\toprule
Sup.        & Weak         & Middle     & Strong          & Model    & Inf. Time & IoU   \\ \midrule
      Full  &              &            &                 & Ours-T   & \textbf{26.08}     & \textbf{56.77}             \\
      10\%  & \checkmark   &            &                 & Baseline & 138.62    & 52.79              \\
      10\%  & \checkmark   &            &                 & Ours-T   & \textbf{26.08}     & 52.66              \\
      10\%  & \checkmark   &            &                 & Ours-S   & \textbf{26.08}     & 52.86               \\
      10\%  & \checkmark   & \checkmark &                 & Ours-S   & \textbf{26.08}     & 52.88               \\
      10\%  & \checkmark   & \checkmark & \checkmark      & Ours-S   & \textbf{26.08}     & 53.60 \\ \bottomrule
\end{tabular}
\end{table}

\begin{table}[H]
\centering                  
\caption{Ablation study of overall pipeline, where ‘Sem.’, ‘Geo.’, and ‘Self.’ denote the semantic branch, geometric branch, and self-supervision in the semantic-geometric fusion module, respectively.}
\label{table6}
\setlength{\tabcolsep}{1mm} 
\begin{tabular}{lllll>{\columncolor{gray!30}}ll}
\toprule
Sup. & Sem. & Geo. & Proj. & Model & mIoU & IoU \\
      \midrule
      Full  & \checkmark & \checkmark  & \checkmark & Ours     & \textbf{20.32}   &\textbf{51.73} \\
      10\%  &            &             &            & Baseline & 9.64    &22.30 \\
      10\%  &            &             & \checkmark & Ours     & 0.37    &7.84 \\
      10\%  & \checkmark &             & \checkmark & Ours     & 11.84   &27.46 \\
      10\%  &            & \checkmark  &  \checkmark& Ours     & 8.08    &17.52 \\
      10\%  & \checkmark & \checkmark  &            & Ours     & 11.55   &18.52 \\
      10\%  & \checkmark & \checkmark  & \checkmark & Ours     & 16.59   &40.60 \\ \bottomrule
\end{tabular}
\end{table}

\subsubsection{Voxel Annotation Ratio}
\label{sec4.4.4}
As shown in Table~\ref{table7}, we use OccFusion~\cite{ming2024occfusion} as the C+L baseline to compare performance under different annotation ratios. OccLE achieves 18.40\% mIoU with only 15\% of voxel annotations, close to our full-data performance (20.32\%). In contrast, the baseline obtains only 6.29\% mIoU under the same setting, much lower than its 100\% performance (21.92\%). These results highlight the effectiveness of our approach in label-efficient scenarios. Our method performs slightly worse than the baseline under full supervision, mainly due to its design for efficiency (inference time 179.5~ms vs. 311.3~ms). Increasing the annotation ratio (e.g., 20\% or 50\%) further narrows the gap with full-data performance. Even with just 5\% of training data, OccLE reaches 13.11\% mIoU, demonstrating strong performance under limited supervision. Additional results under extremely low annotation ratios (1\% or 2\%) are provided in the Appendix \textcolor{cvprblue}{D.5 Part 4}.

\begin{table}[H]
\centering
\setlength{\tabcolsep}{1mm} 
\caption{Ablation study on different voxel annotation ratios. Gray rows: mIoU. White rows: IoU.}
\label{table7}
\begin{tabular}{lllll}
\toprule
Ratio & 5\% & 10\% &15\% & 100\% \\ 
\midrule
Ours      & 32.88 & 40.60 & 42.10 & \textbf{51.73} \\
Baseline \cite{ming2024occfusion}  & 28.56 & 30.36 & 30.37 & \textbf{58.68} \\ 
\midrule
\rowcolor{gray!30}
Ours      & 13.11 & 16.59 & 18.40 & \textbf{20.32} \\
\rowcolor{gray!30}
Baseline \cite{ming2024occfusion}  &  5.16 &  6.03 &  6.29 & \textbf{21.92} \\
\bottomrule
\end{tabular}
\end{table}

\section{Conclusion}
\label{sec5}
In this paper, we propose OccLE, a label-efficient 3D semantic occupancy prediction that leverages limited voxel annotations while preserving high performance. OccLE incorporates three key components: (1) distilling 2D foundation model to predict aligned pseudo label for supervising 2D and 3D semantic learning, (2) cross-plane image and LiDAR feature synergy for efficient geometry learning under semi-supervision, and (3) semantic-geometric feature grid fusion with Dual Mamba and scatter-accumulated projection for supervising unannotated regions. Experiments show that OccLE achieves competitive performance with only 10\% of voxel annotations on the SemanticKITTI and Occ3D-nuScenes datasets.

{
    \small
    \bibliographystyle{ieeenat_fullname}
    \bibliography{main}
}

\input{sec/X_suppl}
\end{document}

%% file: sec/X_suppl.tex
\clearpage
\setcounter{page}{1}
\maketitlesupplementary

\setcounter{section}{0}  

\setcounter{table}{0}
\renewcommand{\thetable}{A\arabic{table}}

\setcounter{figure}{0}
\renewcommand{\thefigure}{A\arabic{figure}}

\setcounter{secnumdepth}{2}
\renewcommand\thesection{\Alph{section}}      
\renewcommand\thesubsection{\thesection.\arabic{subsection}}  

\section{Dual Mamba Structure}
\label{AppA}
The detailed structure of Dual Mamba is illustrated in Fig. \ref{fig5}. The model employs a U-Net backbone with four downsampling stages and four upsampling stages. Each stage incorporates a dual Mamba block. Within a dual Mamba block, two parallel branches compute positional embeddings for the two inputs and then exchange their feature channels prior to reordering. We use a Hilbert curve \cite{hilbert2013dritter} to linearize the 3D representation into a 1D sequence, apply a Mamba block to capture long-range dependencies, and finally restore the three-dimensional structure.
\renewcommand{\thefigure}{A\arabic{figure}}
\setcounter{figure}{0}
\begin{figure}[!htbp] 
\begin{center}
\includegraphics[width=\linewidth]{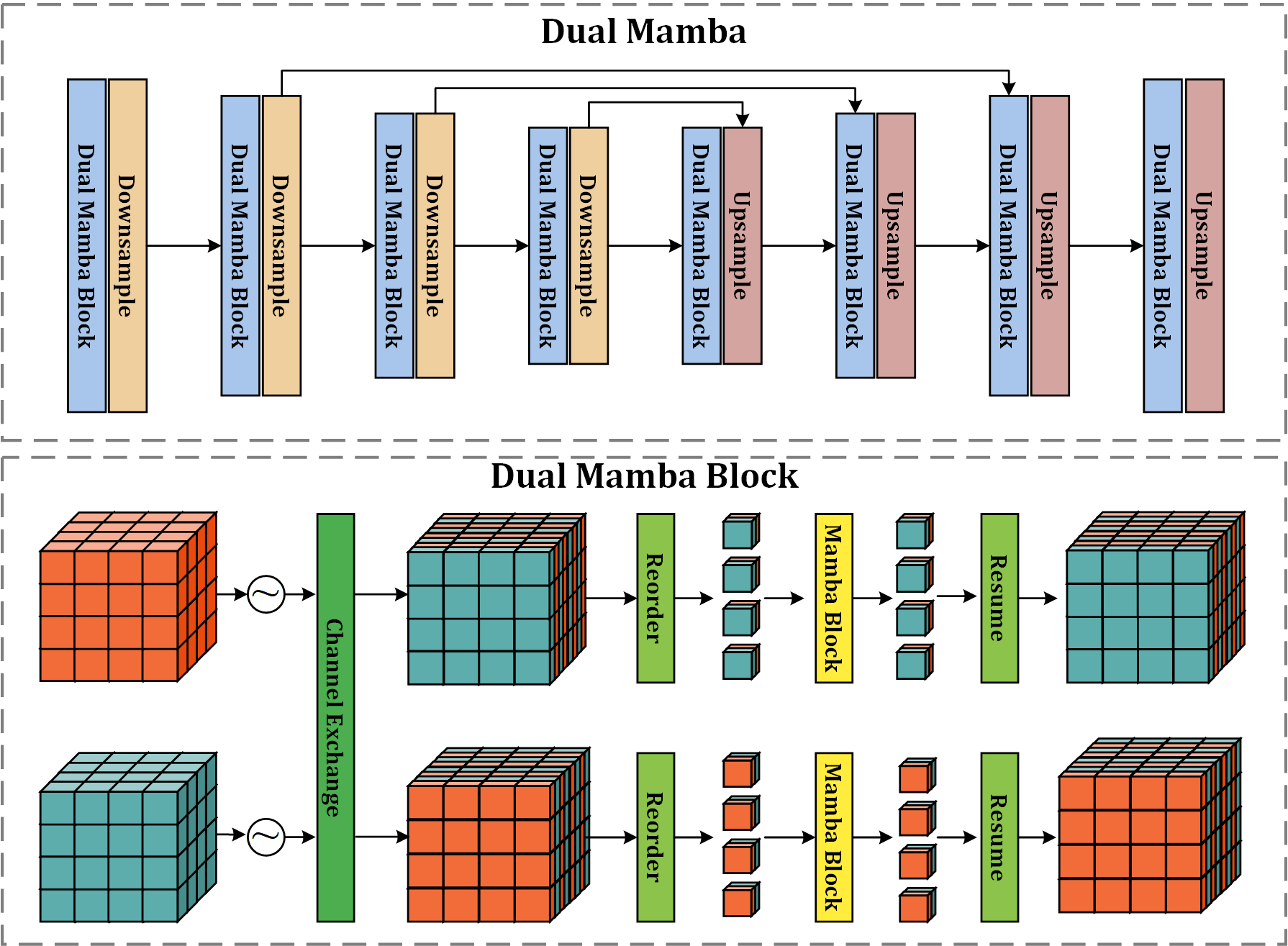}  
\end{center}
\caption{The detailed structure of Dual Mamba. It comprises four stages and employs two parallel branches to process the inputs at each stage.}
\label{fig5}
\end{figure}

\section{Aligned Label Generation}
\label{AppB}
We employ dataset-specific and open-vocabulary segmentation 2D foundation models to generate aligned pseudo labels. The class mappings are listed in Table \ref{tableA1} and Table \ref{tableA2}. We utilize SAM2 \cite{ravi2024sam} to produce trunk class labels and integrate them into the MSeg \cite{lambert2020mseg} outputs. This example illustrates our aligned pseudo label generation process, which can scale to any dataset with different class taxonomies.

\renewcommand{\thetable}{\thesection.\arabic{table}}
\setcounter{table}{0}
\begin{table}[!htb]
\setlength{\tabcolsep}{3pt} 
\centering
\caption{The aligned label generation on the SemanticKITTI dataset.}
\resizebox{\linewidth}{!}{%
\label{tableA1}
\small
\begin{tabular}{llll|llll}   
\toprule
\multicolumn{2}{l}{SemantiKITTI} & \multicolumn{2}{l}{Image Segmentation} & \multicolumn{2}{l}{SemantiKITTI} & \multicolumn{2}{l}{Image Segmentation} \\
Label & Class & Label & Class & Label & Class & Label & Class \\
\midrule
0 & free &  &  & 12 & other-grnd. & 31 & road\_barrier \\
1 & car & 176 & car &  &  & 32 & mailbox \\
2 & bicycle & 175 & bicycle &  &  & 137 & fire\_hydrant \\
3 & motorcycle & 178 & motorcycle &  &  & 191 & wall \\
 &  &  &  & 13 & building & 35 & building \\
4 & truck & 182 & truck & 14 & fence & 144 & fence \\
5 & other.-veh. & 177 & autorickshaw & 15 & vegetation & 131 & road\_barrier \\
 &  & 180 & bus &  &  & 174 & vegetation \\
 &  & 181 & train & 16 & trunk & SAM2 & trunk  \\
 &  & 183 & trailer & 17 & terrain & 102 & terrain \\
 &  & 185 & slow\_wheeled\_object & 18 & pole & 143 & pole \\
6 & person & 125 & person &  &  & 130 & streetlight \\
7 & bicyclist & 126 & rider\_other &  &  & 145 & railing\_banister \\
 &  & 127 & bicyclist &  &  & 146 & guard\_rail \\
8 & motorcyclist & 128 & motorcyclist &  &  & 162 & column \\
9 & road & 98 & road & 19 & traf.-sign. & 135 & traffic\_sign \\
10 & parking & 138 & parking\_meter &  &  & 136 & traffic\_light \\
11 & sidewalk & 100 & sidewalk\_pavement &  &  &  &  \\
\bottomrule
\end{tabular}
}
\end{table}

\begin{table}[!htb]
\setlength{\tabcolsep}{3pt} 
\centering
\caption{The aligned label generation on the Occ3D-nuScenes dataset.}
\resizebox{\linewidth}{!}{%
\label{tableA2}
\begin{tabular}{llll|llll}   
\toprule
\multicolumn{2}{l}{Occ3D-nuScenes} & \multicolumn{2}{l}{Image Segmentation} & \multicolumn{2}{l}{Occ3D-nuScenes} & \multicolumn{2}{l}{Image Segmentation} \\
Label & Class & Label & Class & Label & Class & Label & Class \\
\midrule
0 &other       &rest  &sky et.al           &8 &traffic cone& SAM2 &traffic cone \\
1 & barrier    & 130 & streetlight         & 9 &trailer     & 183 &trailer     \\ 
  &            & 131 & road barrier        & 10 &truck       & 182 &trailer     \\
  &            & 144 & fence               & 11&drive. surf.& 98 &road     \\
  &            & 145 & railing banister    & 12&other flat  & 96 &playing field      \\
  &            & 145 & guard rail          & &            & 97 &railroad      \\
2 & bicycle    & 175 & bicycle             & 13&sidewalk    & 100 &sidewalk pavemen     \\
3 & bus        & 180 & bus                 &14&terrain     & 102 &terrain     \\
4 & car        & 176 & car                 & 15&manmade     &191  &wall       \\
5 &cons. veh   & SAM2 & cons. veh          &  &            &192  &window      \\
6 &motorcycle  & 178 & motorcycle          & &            &132-141&traffic sign et al.    \\
  &  & 128 & motorcyclist                  & 16&vegetation  &174  &vegetation      \\
7 &pedestrian  & 125 & person              & &        & &      \\

\bottomrule
\end{tabular}
}
\end{table}

\section{Training Strategy}
\label{AppC}
The training strategy of OccLE is summarized in Algorithm \ref{alg1}. In Phases I and II, we independently train the semantic and geometric branches. In Phase III, all modules are jointly optimized while freezing the geometric branch to prevent its overfitting.

\begin{algorithm}[ht]
\footnotesize
\caption{The training strategy of OccLE}
\label{alg1}
\begin{algorithmic}[1]
\REQUIRE Annotated dataset $\mathcal{D}_l$, unannotated dataset $\mathcal{D}_u$, semantic branch $\Theta_{\text{sem}}$, geometric branch $\Theta_{\text{geo}}$, fusion module $\Theta_{\text{fus}}$
\ENSURE Trained $\Theta_{\text{sem}}, \Theta_{\text{geo}}, \Theta_{\text{fus}}$

\STATE \textbf{Phase I: Train the semantic branch}

Train $\Theta_{\text{sem}}$ with $\mathcal{D}_l$, $\mathcal{D}_u$, loss $\ell_{\text{sem}}$, label $\{\bar{\mathbf{S}}, \bar{\mathbf{s}}\}$

\STATE \textbf{Phase II: Train the geometric branch}

Train teacher model $\Theta_{\text{geo-T}}$ with $\mathcal{D}_l$, loss $\ell_{\text{sem}}$, label $\{\bar{\mathbf{G}}\}$

Predict $\{\bar{\mathbf{G}}_T\}$ for $\mathcal{D}_u$ via $\Theta_{\text{geo-T}}$

Train student model $\Theta_{\text{geo-S}}$ with $\mathcal{D}_l$, $\mathcal{D}_u$, loss $\ell_{\text{sem}}$, labels $\{\bar{\mathbf{G}}\}$, $\{\bar{\mathbf{G}}_T\}$

\STATE \textbf{Phase III: Train the fusion module}

Freeze $\Theta_{\text{geo-S}}$

Train $\Theta_{\text{fus}}$ and $\Theta_{\text{sem}}$ with $\mathcal{D}_l$, $\mathcal{D}_u$, losses $\ell_{\text{fus}}$, $\ell_{\text{sem}}$, labels $\{\bar{\mathbf{Y}}, \bar{\mathbf{s}}\}$
\end{algorithmic}
\end{algorithm}

\section{Supplementary Experiments}
\label{AppD}

\begin{table*}[!htb]
\centering
\caption{Quantitative results on SemanticKITTI hidden test set. \textbf{Bold} and \underline{underline} represent the best and second best results, respectively. Inp. and Sup. indicate the input modality and the supervision type, respectively.}
\label{tableA3}
\setlength{\tabcolsep}{1mm} 
\resizebox{\linewidth}{!}{%
\begin{tabular}{l|l|l|l>{\columncolor{gray!30}}l|lllllllllllllllllll}
\toprule
Method &Inp.&Sup. & IoU & mIoU &\rotatebox{90}{{\tikz \fill[c1] (0,0) rectangle (0.6em,0.6em);} car (3.92\%)} & \rotatebox{90}{{\tikz \fill[c2] (0,0) rectangle (0.6em,0.6em);} bicycle (0.03\%)} & \rotatebox{90}{{\tikz \fill[c3] (0,0) rectangle (0.6em,0.6em);} motorcycle (0.03\%)} & \rotatebox{90}{{\tikz \fill[c4] (0,0) rectangle (0.6em,0.6em);} truck (0.16\%)} & \rotatebox{90}{{\tikz \fill[c5] (0,0) rectangle (0.6em,0.6em);} other-veh. (0.20\%)} & \rotatebox{90}{{\tikz \fill[c6] (0,0) rectangle (0.6em,0.6em);} person (0.07\%)} & \rotatebox{90}{{\tikz \fill[c7] (0,0) rectangle (0.6em,0.6em);} bicyclist (0.07\%)} & \rotatebox{90}{{\tikz \fill[c8] (0,0) rectangle (0.6em,0.6em);} motorcyclist (0.05\%)} & \rotatebox{90}{{\tikz \fill[c9] (0,0) rectangle (0.6em,0.6em);} road (15.30\%)} & \rotatebox{90}{{\tikz \fill[c10] (0,0) rectangle (0.6em,0.6em);} parking (1.12\%)} & \rotatebox{90}{{\tikz \fill[c11] (0,0) rectangle (0.6em,0.6em);} sidewalk (11.13\%)} & \rotatebox{90}{{\tikz \fill[c12] (0,0) rectangle (0.6em,0.6em);} other-grnd.(0.56\%)} & \rotatebox{90}{{\tikz \fill[c13] (0,0) rectangle (0.6em,0.6em);} building (14.10\%)} & \rotatebox{90}{{\tikz \fill[c14] (0,0) rectangle (0.6em,0.6em);} fence (3.90\%)} & \rotatebox{90}{{\tikz \fill[c15] (0,0) rectangle (0.6em,0.6em);} vegetation (39.3\%)} & \rotatebox{90}{{\tikz \fill[c16] (0,0) rectangle (0.6em,0.6em);} trunk (0.51\%)} & \rotatebox{90}{{\tikz \fill[c17] (0,0) rectangle (0.6em,0.6em);} terrain (9.17\%)} & \rotatebox{90}{{\tikz \fill[c18] (0,0) rectangle (0.6em,0.6em);} pole (0.29\%)} & \rotatebox{90}{{\tikz \fill[c19] (0,0) rectangle (0.6em,0.6em);} traf.-sign (0.08\%)} \\ 
\midrule
TPVFormer \cite{huang2023tri} & C & Full & 34.25 & 11.26 & 19.20 & 1.00 & 0.50 & 3.70 & 2.30 & 1.10 & 2.40 & 0.30 & 55.10 & 27.40 & 27.20 & 6.50 & 14.80 & 11.00 & 13.90 & 2.60 & 20.40 & 2.90 & 1.50 \\
OccFormer \cite{zhang2023occformer} & C & Full  & 34.53 & 12.32 & 21.30 & 1.50 & 1.70 & 3.90 & 3.20 & 2.20 & 1.10 & 0.20 & 55.90 & \underline{31.50} & 30.30 & 6.50 & 15.70 & 11.90 & 16.80 & 3.90 & 21.30 & 3.80 & 3.70 \\
VoxFormer \cite{li2023voxformer}& C & Full  & 43.21 & 13.41 & 21.70 & 1.90 & 1.60 & 3.60 & 4.10 & 1.60 & 1.10 & 0.00 & 54.10 & 25.10 & 26.90 & 7.30 & 23.50 & 13.10 & 24.40 & 8.10 & 24.20 & 6.60 & 5.70 \\
SurroundOcc \cite{wei2023surroundocc} & C & Full  & 34.72 & 11.86 & 20.30 & 1.60 & 1.20 & 1.40 & 4.40 & 1.40 & 2.00 & 0.10 & 56.90 & 30.20 & 28.30 & 6.80 & 15.20 & 11.30 & 14.90 & 3.40 & 19.30 & 3.90 & 2.40 \\
H2GFormer \cite{wang2024h2gformer} & C & Full  & 43.52 & 14.60 & 23.70 & 0.60 & 1.20 & \underline{5.20} & 5.00 & 1.10 & 0.10 & 0.00 & 57.90 & 30.00 & 30.40 & 6.90 & 24.00 & 14.60 & 25.20 & 10.70 & 25.80 & 7.50 & 7.10 \\
SGN \cite{mei2024camera}  & C & Full  & \textbf{45.42} & 15.76 & \underline{25.40} & \textbf{4.50} & 0.90 & 4.50 & 3.70 & 0.50 & 0.30 & 0.10 & \underline{60.40} & 28.90 & \underline{31.40} & 8.70 & \textbf{28.40} & 18.10 & \underline{27.40} & \underline{12.60} & \underline{28.40} & \underline{10.00} & 8.30 \\
LowRankOcc \cite{zhao2024lowrankocc} & C & Full  & 38.47 & 13.56 & 20.90 & 3.30 & \underline{2.70} & 2.90 & 4.40 & \underline{2.40} & 1.70 & \textbf{2.30} & 52.80 & 25.10 & 27.20 & 8.80 & 22.10 & 14.40 & 22.90 & 8.90 & 20.80 & 7.00 & 7.00 \\
Symphonies \cite{jiang2024symphonize} & C & Full  & 42.19 & 15.04 & 23.60 & \underline{3.60} & 2.60 & 3.20 &5.60 & \textbf{3.20} & 1.90 & \underline{2.00} & 58.40 & 26.90 & 29.30 & \textbf{11.70} & 24.70 & 16.10 & 24.20 & 10.00 & 23.10 & 7.70 & 8.00 \\
HASSC \cite{wang2024not} & C & Full  & 42.87 & 14.38 & 23.00 & 1.90 & 1.50 &2.90 & 4.90 & 1.40 & \underline{3.00} & 0.00 & 55.30 & 25.90 & 29.60 & \underline{11.30} & 23.10 & 14.30 & 24.80 & 9.80 & 26.50 & 7.00 & 7.10 \\
Bi-SSC \cite{xue2024bi} & C & Full  & \underline{45.10} & \underline{16.73} & 25.00 & 1.80 & \textbf{2.90} & \textbf{6.80} & \textbf{6.80} & 1.70 & \textbf{3.30} & 1.00 & \textbf{63.40} & \textbf{31.70} & \textbf{33.30} & 11.20 & \underline{26.60} & \underline{19.40} & 26.10 & 10.50 & \textbf{28.90} & 9.30 & \underline{8.40} \\ 
\midrule
OccLE (Ours) & C+L & 10\%  & 31.42 & \underline{16.30} & \textbf{29.60} & 3.50 &\underline{2.70} & 0.00& \underline{6.30} & 1.90 & 0.00 & 0.00 & 56.90 & 22.60 & 30.80 & 2.30 & 24.90 &\textbf{21.90} & \textbf{30.00} & \textbf{19.10} &27.90 & \textbf{15.10}&\textbf{14.00}\\
\bottomrule
\end{tabular}
}
\end{table*}

\begin{table*}[!htb]
\centering
\caption{Quantitative results on  SSCBench-KITTI-360 validation set.}
\label{tableA4}
\setlength{\tabcolsep}{1mm} 
\resizebox{\linewidth}{!}{%
\begin{tabular}{l|l|l|l>{\columncolor{gray!30}}l|llllllllllllllllll}
\toprule
Method &Inp.&Sup. & IoU & mIoU &\rotatebox{90}{{\tikz \fill[c1] (0,0) rectangle (0.6em,0.6em);} car (2.85\%)} & \rotatebox{90}{{\tikz \fill[c2] (0,0) rectangle (0.6em,0.6em);} bicycle (0.01\%)} & \rotatebox{90}{{\tikz \fill[c3] (0,0) rectangle (0.6em,0.6em);} motorcycle (0.01\%)} & \rotatebox{90}{{\tikz \fill[c4] (0,0) rectangle (0.6em,0.6em);} truck (0.16\%)} & \rotatebox{90}{{\tikz \fill[c5] (0,0) rectangle (0.6em,0.6em);} other-veh. (5.75\%)} & \rotatebox{90}{{\tikz \fill[c6] (0,0) rectangle (0.6em,0.6em);} person (0.02\%)} &\rotatebox{90}{{\tikz \fill[c9] (0,0) rectangle (0.6em,0.6em);} road (14.98\%)} & \rotatebox{90}{{\tikz \fill[c10] (0,0) rectangle (0.6em,0.6em);} parking (2.31\%)} & \rotatebox{90}{{\tikz \fill[c11] (0,0) rectangle (0.6em,0.6em);} sidewalk (6.43\%)} & \rotatebox{90}{{\tikz \fill[c12] (0,0) rectangle (0.6em,0.6em);} other-grnd.(2.05\%)} & \rotatebox{90}{{\tikz \fill[c13] (0,0) rectangle (0.6em,0.6em);} building (15.67\%)} & \rotatebox{90}{{\tikz \fill[c14] (0,0) rectangle (0.6em,0.6em);} fence (0.96\%)} & \rotatebox{90}{{\tikz \fill[c15] (0,0) rectangle (0.6em,0.6em);} vegetation (41.99\%)}  & \rotatebox{90}{{\tikz \fill[c17] (0,0) rectangle (0.6em,0.6em);} terrain (7.10\%)} & \rotatebox{90}{{\tikz \fill[c18] (0,0) rectangle (0.6em,0.6em);} pole (0.22\%)} & \rotatebox{90}{{\tikz \fill[c19] (0,0) rectangle (0.6em,0.6em);} traf.-sign (0.06\%)} & \rotatebox{90}{{\tikz \fill[c20] (0,0) rectangle (0.6em,0.6em);} other-struct. (4.33\%)}  & \rotatebox{90}{{\tikz \fill[c21] (0,0) rectangle (0.6em,0.6em);} other-obj. (0.28\%)}\\ 
\midrule
TPVFormer \cite{huang2023tri}                   & C & Full & 40.22 & 13.64 & 21.56 & 1.09 & 1.37 & 8.06 & 2.57 & 2.38 & 52.99 & 11.99 & 31.07 & 3.78 & 34.83 & 4.80 & 30.08 & 17.52 & 7.46 & 5.86 & 5.48 & 2.70 \\
OccFormer \cite{zhang2023occformer}             & C & Full & 40.27 & 13.81 & 22.58 & 0.66 & 0.26 & 9.69 & 3.82 & 2.77 & 54.30 & 13.44 & 31.53 & 3.55 &36.42 & 4.80 & 31.00 & 19.51 & 7.77 & 8.51 & 6.95 & 4.60 \\
VoxFormer \cite{li2023voxformer}                & C & Full & 38.76 & 11.91 & 17.84 & 1.16 & 0.89 & 4.56 & 2.06 & 1.63 & 47.01 & 9.67 & 27.21 & 2.89 & 31.18 & 4.97 & 28.99 & 14.69 & 6.51 & 6.92 & 3.79 & 2.43 \\
SGN \cite{mei2024camera}                        & C & Full & 47.06 & \underline{18.25} & 29.03 &\textbf{3.43} & 2.90 & 10.89 & 5.20 & 2.99 & 58.14 & \underline{15.04} & \underline{36.40} & 4.43 & 42.02 & \underline{7.72} & 38.17 & 23.22 & \underline{16.73} & \textbf{16.38} & \underline{9.93} & \underline{5.86} \\
Symphonies \cite{jiang2024symphonize}           & C & Full & 44.12 & \textbf{18.58} & 30.02 & 1.85 & \textbf{5.90} & \textbf{25.07} & \textbf{12.06} & \textbf{8.20} &54.94 & 13.83 & 32.76 & \textbf{6.93} & 35.11 &\textbf{8.58} & 38.33 & 11.52 & 14.01 & 9.57 & \textbf{14.44} & \textbf{11.28} \\
GaussianFormer \cite{huang2025gaussianformer}   & C & Full & 35.38 & 12.92 &18.93 & 1.02 &\underline{4.62} & \underline{18.07} & \underline{7.59} & \underline{3.36} & 45.47 & 10.89 & 25.03 & \underline{5.32} & 28.44 & 5.68 & 29.54 & 8.62 & 2.99 & 2.32 & 9.51 & 5.14 \\
GaussianForme-2 \cite{huang2025gaussianformer2}  & C & Full& 38.37 &13.90 &21.08 &\underline{2.55} &4.21 &12.41 &5.73 &1.59 &54.12 &11.04 &32.31 &3.34 &32.01 &4.98 &28.94 &17.33 &3.57 &5.48 &5.88 &3.54 \\
 LMSCNet\cite{roldao2020lmscnet}               & L & Full &47.53 &13.65 &20.91 &0.00 &0.00 &0.26 &0.00 &0.00 &\underline{62.95}&13.51&33.51 &0.20 &43.67&0.33 &40.01&\underline{26.80} &0.00 &0.00 &3.63 &0.00 \\
 SSCNet\cite{song2017semantic}                 & L & Full &\textbf{53.58} &16.95 & \textbf{31.95} &0.00 &0.17 &10.29&0.58 &0.07 &\textbf{65.70} &\textbf{17.33}&\textbf{41.24} &3.22 &\underline{44.41} &6.77 &\textbf{43.72}&\textbf{28.87} &0.78 &0.75 &8.60 &0.67 \\
\midrule
OccLE (Ours) & C+L & 10\% & \underline{52.44} & 16.38 & \underline{31.14} & 0.34 & 0.28 & 4.66 & 1.39 & 2.05 &53.47 & 9.15 & 29.65 & 4.32 & \textbf{45.07} & 7.30 & \underline{41.53} & 23.79 & \textbf{19.81} & \underline{11.36} & 6.45 & 3.11  \\
\bottomrule
\end{tabular}
}
\end{table*}

\subsection{Supplementary Datasets}
\label{AppD1}
 We evaluate OccLE on additional datasets: SSCBenchKITTI-360 \cite{li2024sscbench}. SSCBenchKITTI-360 shares its scene and voxel configuration with SemanticKITTI and annotates voxels for 19 class labels (18 Semantic + 1 Free). 

\subsection{Supplementary Implementation Details}
\label{AppD2}
In the semantic branch, the downsampling scale of ${{{\cal G}_{s2d}}}$ and ${{{\cal G}_{s3d}}}$  is ${8 \times }$; In the geometric branch, we utilize ResNet50 \cite{he2016deep} and a 2-layer sparse Conv3D to extract features from image and LiDAR scans. In the semantic-geometric feature grid fusion module, we stack 4 Dual Mamba blocks. For the SemanticKITTI dataset, we adopt the multi-frame setting as in \cite{li2023voxformer, mei2024camera, wang2024h2gformer, wang2024not}, and crop the camera images to ${1220 \times 370}$. For the Occ3D-nuScenes dataset, we use a single-frame setting and crop the images to ${1600 \times 900}$. For the SSCBenchKITTI-360 dataset, we follow the same multi-frame setting as SemanticKITTI and crop the images to ${1408 \times 376}$.

\subsection{Supplementary Quantitative Comparison}
\label{AppD3}

The quantitative comparison results on the SemanticKITTI hidden test set and SSCBenchKITTI-360 validation set are presented in Table \ref{tableA3} and Table \ref{tableA4}. OccLE achieves 16.30 mIoU and 31.42 IoU on the SemanticKITTI hidden test set, ranking second among camera-based fully supervised methods. On the SSCBenchKITTI-360 validation set, OccLE achieves 16.38 mIoU and 52.44 IoU, performing competitively against both camera-based and LiDAR-based methods. These results indicate that OccLE can perform well even with limited voxel-level annotations.

\renewcommand{\thetable}{\thesection.\arabic{table}}
\setcounter{table}{1}
\begin{table*}[!htb]
\setlength{\tabcolsep}{3pt} 
\centering
\caption{Efficiency comparison between methods with different representative inputs and supervision. Inf. Time, Inp., and Sup. indicate the inference time, input modality, and supervision type, respectively.}
\label{tableA5}
\begin{tabular}{llllllll}   
\toprule
Method  &Ours &OccFormer \cite{zhang2023occformer} &VoxFormer \cite{li2023voxformer} &SGN \cite{mei2024camera} &Symphonies \cite{jiang2024symphonize} &OccFusion \cite{ming2024occfusion} &SelfOcc \cite{huang2024selfocc} \\
\midrule
Inp. &C+L  &C & C & C & C & C+L & C \\
Sup. & 10\%&Full & Full & Full & Full & Full & Self \\
Inf. Time &179.5 & 311.3 &116.1 &243.7 &\textbf{105.0} &198.7 &200.0\\
\bottomrule
\end{tabular}
\end{table*}

\subsection{Efficiency Comparison}
\label{AppD4}
We report the inference time on a single A6000 Ada GPU for methods with different representative inputs and supervision in Table \ref{tableA5}. The experimental results demonstrate the superiority of OccLE compared with the C+L input method OccFusion \cite{ming2024occfusion} (179.5 ms vs. 198.7 ms), benefiting from several efficiency-oriented design choices such as the geometry branch. Moreover, OccLE still shows competitive performance compared to methods using only the C input.

\subsection{Supplementary Ablation Study}
\label{AppD5}

\textbf{Semantic Alignment Failure.}
\label{AppD5.1} 
To evaluate the effect of semantic alignment on the semantic branch, we examine two cases: one without semantic alignment and another that simulates potential alignment errors by adding noise to the semantic segmentation map during the training phase. As shown in Table \ref{tableA6}, only using Mseg \cite{lambert2020mseg} results as supervision causes some categories to completely fail under effective supervision, reducing the mIoU to 15.80. When a certain proportion of noise is added to the training samples, the model performance slightly decreases but still maintains a high mIoU.

\begin{table}[H]
\setlength{\tabcolsep}{3pt} 
\centering
\caption{Ablation study on semantic alignment. The semantic noise ratio is defined relative to the number of pixels.}
\label{tableA6}
\begin{tabular}{lllll}   
\toprule 
                & Baseline&  w/o Align. &	w/ 2\% Noise	& w/ 6\% Noise \\\midrule
\rowcolor{gray!30} mIoU& \textbf{16.59} & 15.80	      & 16.01	    & 15.37\\
\bottomrule
\end{tabular}
\end{table}

\noindent \textbf{Geometry Learning Failure.}
\label{AppD5.2} 
To evaluate the potential modality failure in the geometry branch, we apply the student model's augmentation strategy from the training phase to the inference inputs to simulate geometry learning failure. As shown in Table \ref{tableA7}, image degradation has the most significant impact on mIoU in the fusion stage, while point dropping most strongly affects IoU in the geometry branch. This observation is consistent with the modality characteristics discussed in Sec. \textcolor{cvprblue}{3.3}.

\begin{table}[H]
\setlength{\tabcolsep}{3pt} 
\centering
\caption{Ablation study of geometry learning failure. ID and PD denote image degradation and point dropping, respectively.}
\label{tableA7}
\begin{tabular}{lllll}   
\toprule 
Fail. Case     &Baseline &  ID   & PD	  & ID+PD \\\midrule
IoU (Geo.)	   & \textbf{53.60}  &52.88  &	52.87 &	52.79 \\
\rowcolor{gray!30} mIoU (Fus.)&\textbf{16.59}	&14.49  &	15.95 &	14.1 \\
\bottomrule
\end{tabular}
\end{table}

\noindent \textbf{Pipeline and Component.}
\label{AppD5.3}
To assess the impact of semi-supervision in the geometric branch on the overall pipeline, As shown in Table \ref{tableA8}, we incorporated a teacher model of the geometric branch into final joint training. The results indicate that mIoU and IoU decrease by 0.42 and 1.74, respectively. We compare our scatter-accumulated projection method with an alternative approach that weights features by the distance between each voxel and the camera. The latter method yields a 3.21 mIoU drop. This result demonstrates that the scatter-accumulated projection is simple yet effective in supervising all voxels with aligned pseudo labels, whereas the distance-weighted projection reduces supervision quality for distant voxels. 

\begin{table}[H]
\setlength{\tabcolsep}{3pt} 
\centering
\caption{Ablation study of pipeline and component.}
\label{tableA8}
\begin{tabular}{l>{\columncolor{gray!30}}ll}   
\toprule
Model & mIoU & IoU \\
\midrule
w/ Geometric Teacher Model	&16.17	&   38.86\\
w/ Weighted Projection      &13.38  &	40.40\\
Ours                        &\textbf{16.59}  &	\textbf{40.60}\\
\bottomrule
\end{tabular}
\end{table}

\noindent \textbf{Extreme Low Voxel Annotation Ratios}
\label{AppD5.4}
To evaluate OccLE performance under extreme low voxel annotation ratios, we train OccLE using only 1\% and 2\% of voxel annotations. In the 1\% case, this corresponds to only 30 training samples. As shown in Table \ref{tableA9}, despite such limited supervision, the model still achieves an mIoU of around 10\%, demonstrating remarkable label efficiency.

\begin{table}[H]
\centering
\setlength{\tabcolsep}{1mm} 
\caption{Ablation study on extreme low voxel annotation ratios.}
\label{tableA9}
\begin{tabular}{ll>{\columncolor{gray!30}}l}
\toprule
Ratio & IoU & mIoU \\ \midrule
2\%  & 21.11 & 10.97 \\
1\%  & 18.15 & 9.70 \\ 
\bottomrule
\end{tabular}
\end{table}

\subsection{Supplementary Qualitative Comparison}
\label{AppD6}
We show additional qualitative results of OccLE on the SemanticKITTI validation dataset in Figure \ref{fig6}. OccLE maintains strong performance across varied scenes, delivering effective scene completion and precise class classification.
\renewcommand{\thefigure}{C\arabic{figure}}
\setcounter{figure}{0}
\begin{figure*}[!htbp] 
\begin{center}
\includegraphics[width=\linewidth]{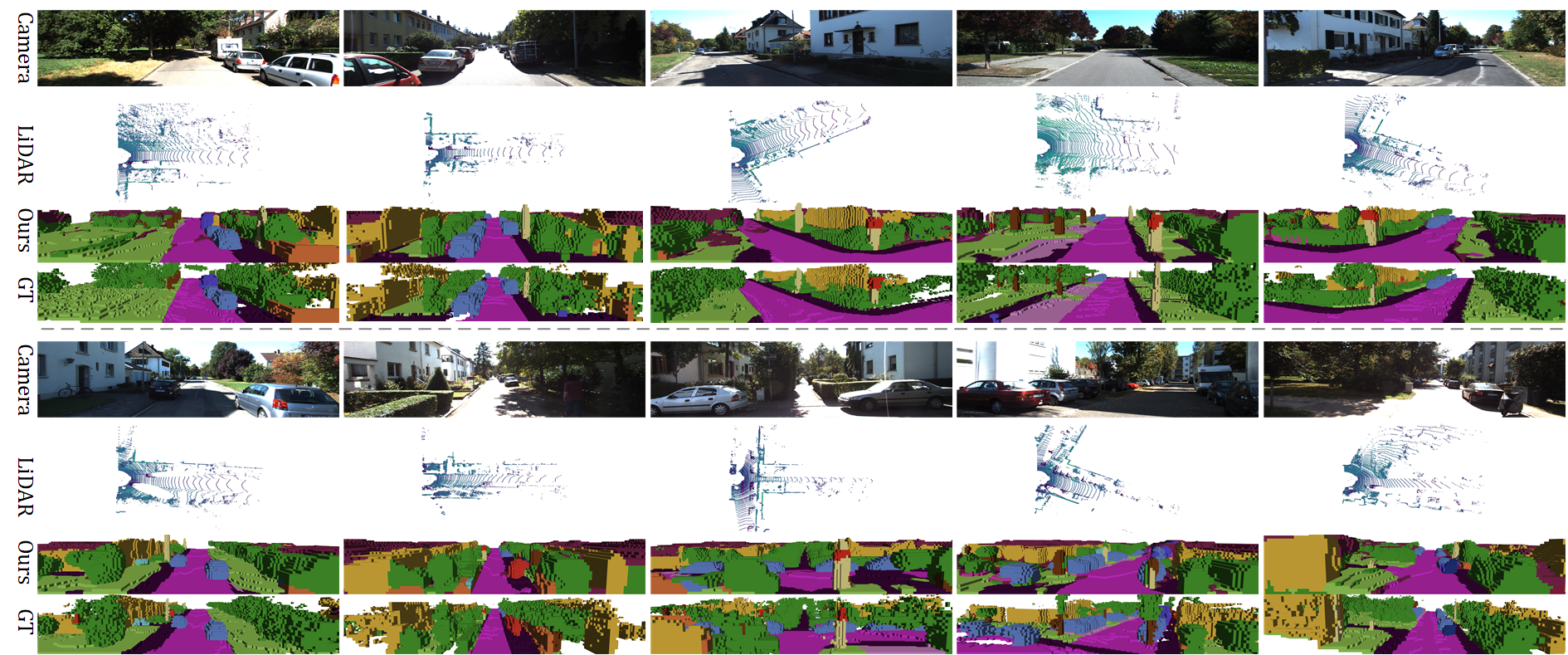}  
\end{center}
\caption{The qualitative results of OccLE on SemanticKITTI validation dataset.}
\label{fig6}
\end{figure*}

\section{Limitation}
\label{AppE}
In this study, we simulate label-efficient learning for 3D semantic occupancy prediction by uniformly sampling 10\% of voxel annotations across all scenes in the SemanticKITTI dataset. This strategy ensures the generalization across diverse scenes. However, concentrating limited voxel annotations within a few scenes may lead to overfitting, thereby diminishing the robustness of OccLE in unfamiliar scenes. Consequently, for real-world autonomous driving applications, it is advisable to distribute limited voxel annotations across a wide range of scenes, enhancing its performance in varied and unseen environments.